\documentclass[final,5p,times,twocolumn]{elsarticle}

\usepackage{amssymb}
\usepackage{amsmath}
\usepackage{graphicx}
\usepackage{xcolor}
\usepackage{booktabs}
\usepackage{array}
\usepackage{pifont}
\usepackage{algorithm}
\usepackage{algpseudocode}
\usepackage{xurl}

\newtheorem{proposition}{Proposition}
\newproof{proof}{Proof}

\newif\ifsubmission
\submissionfalse

\newif\ifclaimmarks
\ifsubmission\claimmarksfalse\else\claimmarkstrue\fi
\newcommand{\HAclaimmark}[3]{%
  \ifclaimmarks
    {\color{#1}#3}%
    \if\relax\detokenize{#2}\relax\else
      \textsuperscript{\color{#1}[#2]}%
    \fi
  \else
    #3%
  \fi}

\begin{document}

\begin{frontmatter}

\title{When to Trust the BIM: Identifiability-Aware LiDAR-Inertial--BIM Fusion under As-Built Deviations}

\author[a]{Yujie Zhang}
\author[a]{Yuxuan Guo}
\author[a]{Xinbao Sun}
\author[a]{Chunlei Zhao}
\author[a]{Langqi Miao}
\author[a]{Xiaohui Jia\corref{cor1}}
\ead{2010081@hebut.edu.cn}
\author[a]{Jinyue Liu\corref{cor2}}
\ead{lyj@hebut.edu.cn}

\cortext[cor1]{Corresponding author.}
\cortext[cor2]{Co-corresponding author.}

\affiliation[a]{organization={School of Mechanical Engineering, Hebei University of Technology},
            addressline={No. 5340 Xiping Road, Beichen District},
            city={Tianjin},
            postcode={300401},
            country={China}}

\begin{abstract}
Building information models (BIM) can provide a global geometric reference for robot localization with LiDAR-inertial odometry (LIO). However, as-built deviations from the as-designed model mean that a disagreement between a scan and the BIM may arise from a pose error, an offset wall surface (an as-built deviation or an unmodeled object in front of the wall), or a shifted room, and in some geometries these explanations produce identical observations. In this paper, we formalize this attribution problem as fault identification over three fault classes and derive a geometric criterion for when attribution is possible. Building on it, we propose an identifiability-aware method that reports an attribution only when the fault is attributable; otherwise, it retains the candidates, defers BIM updates and BIM-based pose corrections in the ambiguous directions, and recovers once new references arrive. In the pre-registered two-dimensional study, on cases whose fault is detectable but not attributable, the method's conditional error rate was 0.83\%, against 31.67\% for a baseline with uncertainty-based abstention at the same answer rate. On the 5,600 attributable cases, the method reported a unique attribution 5,574 times, with no incorrect attributions. In paired fault injections in a laboratory room with its IFC model and on three floors of the SLABIM dataset, it raised no off-target attribution (0 of 168 conditions in the laboratory; 0 of 1,302 on SLABIM) and attributed all eight laboratory faults predicted to be attributable, whereas both reproduced baselines raised off-target structural flags in every setting.
\end{abstract}


\end{frontmatter}

\section{Introduction}
\label{sec:intro}

BIM is increasingly used as a global geometric reference for robot localization with LiDAR-inertial odometry. What a BIM records, however, is the building as it was designed, and the building as constructed can deviate from that design. When a LiDAR scan disagrees with the model, a robot that takes its global reference from the BIM~\cite{Stuehrenberg2025LIOBIM, Wang2026iBIMLIO, Zhang2026BIMLoc} therefore has to decide which of three things moved: its own pose, the surface at one wall, or a whole room. In some layouts the scan looks the same in all three cases.

\begin{figure}[t]
\centering
\includegraphics[width=\linewidth]{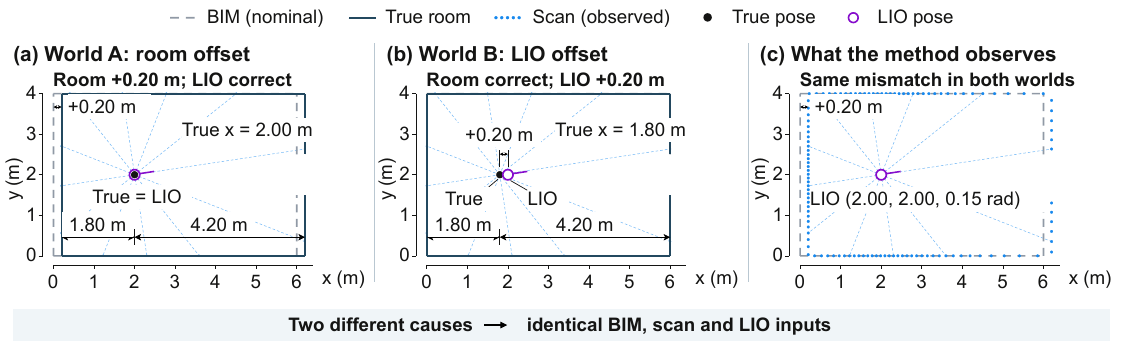}
\caption{Paired worlds. (a) World A: the as-built room is offset by $d = 0.20$~m from its BIM position and the LiDAR-inertial odometry (LIO) pose is correct. (b) World B: the room matches the BIM and the LIO pose is offset by $d$. (c) Both worlds give the same scan and the same LIO pose, so no method can be right in both.}
\label{fig:paired-worlds}
\end{figure}

Consider a robot that sees a single room (Fig.~\ref{fig:paired-worlds}). In world A, the room was built 20~cm away from its position in the BIM, and the LIO pose is correct; in world B, the room matches the BIM, and the LIO pose is 20~cm off. The two worlds give the same scan and the same LIO pose, so no method can tell them apart from these data or be right in both. We call such a pair paired worlds. This ambiguity matters because scans are already used to update the BIM toward the as-built state~\cite{Wang2025InformativeAsBuilt, Spinner2024OnlineBIM}, and a wrong revision pollutes that model. A pose pulled toward an unverified BIM sends navigation to the wrong place and then hides the deviation once the map is laid over the BIM.

Even so, current LiDAR--BIM systems return one answer on this pair. Localization against a floor plan or a BIM treats the model as the reference~\cite{Boniardi2019CADLocalization, Li2020ImpreciseMaps, Gao2022FPLoc, Yin2023SemanticBIM}. LIO-BIM adds a BIM factor only after the scan-to-BIM match passes a quality test~\cite{Stuehrenberg2025LIOBIM}, and BIM-Loc admits only structure it has judged consistent~\cite{Zhang2026BIMLoc}. diS-Graphs estimates wall and room deviations in stages, starting from an alignment on walls without deviation, and its authors show that a single joint optimization confuses pose error with structural error~\cite{Shaheer2025DiSGraphs}. These methods answer three different questions: where the robot is, whether the match is good enough to use, and how large the deviations are once an alignment has been fixed. Each method returns one pose, and diS-Graphs also returns one set of deviations, so in the paired worlds, each can be right in at most one world. Reporting a covariance with that pose does not close the gap, because a covariance computed on the assumption that the map is correct cannot reflect an error in the map. Localization, gating and staged estimation take the BIM as the reference they measure against; the fusion below treats it as one more source of evidence and checks that evidence whenever the scan disagrees with it.

Before a conclusion is reported, or the pose is corrected with the BIM, the explanations have to be tested, because they are not always separable. An LIO bias offsets the absolute pose in $x$, $y$ and yaw. A wall-surface offset moves the surface observed at one BIM wall along its normal, and a room translation shifts a whole room in the plane relative to the coordinates at which the run started. The test is decided by the witnesses, that is, the walls and references that independently constrain the offset direction. In the single room the walls of the room together count as one witness against the prior, so those two witnesses detect a conflict, but a pure translation still fits a room translation and an LIO bias alike. In this single-room translation, the cause can be attributed only once the wall outside the door, the third witness, is seen. Whether the explanations can be told apart is set by the geometry of those witnesses, and more data in the same geometry does not change it. Where the geometry separates them, whether a given stretch of the trajectory can tell them apart also depends on the size of the offset relative to the noise. When applied to the fault classes defined by a BIM, this geometric test is the identifiability criterion of the detection--identification--adaptation (DIA) framework~\cite{Teunissen2018DIA, Zaminpardaz2019Identifiability}.

In addition, LiDAR-inertial odometry is kept free of the BIM, so it remains an independent witness. At each conflict, every single-fault explanation is fitted. Any explanation that leaves part of the data unexplained, leaving it for another explanation to account for, is dropped. If the cause can be uniquely attributed, the attribution is reported and the pose is corrected from it. Where several remain, the candidates are kept and the pose follows the odometry wherever they disagree on a BIM-anchored position, until a new witness separates them. Here, we call those directions ambiguous, and the rule deferred BIM anchoring. If the LIO is wrong, the error in an ambiguous direction is about the difference between the candidates until the cause is attributed. With no explanation left, the position is withdrawn.

In the framework of this paper, withholding an attribution is allowed when the witnesses cannot separate the explanations, and fewer answers then mean fewer recorded errors. The errors are therefore compared at the same answer rate, an answer being a reported attribution to a wall, a room, or the LIO. Because the specification was fixed before the runs, this is a pre-registered simulation. Its baseline is a diS-Graphs-style reproduction of~\cite{Shaheer2025DiSGraphs} that withholds an answer where its own covariance is large. Where the noise model labels a fault detectable but not attributable, the method almost always withholds an attribution.

On the 7,200 simulated cases with that label, the conditional error rate at the same answer rate was 0.83\% (1 of 120 answers) for our method, against 31.67\% (38 of 120) for the abstaining baseline. The difference meets the pre-registered criterion for a clear improvement. On the 5,600 attributable cases, our method reported a unique attribution 5,574 times and made no wrong attribution. In a separate demonstration with a strong prior, a new reference resolved the paired worlds in all 200 room-world trials and in 194--195 of 200 LIO-world trials, with no wrong unique attribution. Our method raised no off-target attribution in 168 injected conditions in a laboratory room with its IFC model, nor in 1,302 conditions on SLABIM~\cite{Huang2025SLABIM}, and it attributed all eight laboratory faults predicted to be attributable.

The main contributions are summarized as follows:

\begin{enumerate}
\item \textbf{Identifiability criterion.} A geometric criterion for when a scan--BIM conflict can be uniquely attributed to an LIO bias, a wall-surface offset, or a room translation. The criterion follows from the observation that a narrower explanation contained in a broader one can never be confirmed on its own; in a single room, a room translation and an LIO bias form such a pair (Section~\ref{sec:identifiability}).
\item \textbf{Deferred BIM anchoring.} A fusion method that reports an attribution only when the cause can be uniquely attributed, and that otherwise retains the candidates and defers BIM anchoring in the ambiguous directions (Section~\ref{sec:method}).
\item \textbf{Evaluation under known faults.} An evaluation that compares errors at a fixed answer rate in pre-registered simulation, together with paired fault injections on a laboratory room and on three floors of SLABIM, where the ground-truth fault is known because it was inserted (Sections~\ref{sec:design} and~\ref{sec:results}).
\end{enumerate}

The criterion answers the question in the title: the BIM can be trusted where the geometry makes the fault attributable and the offset is large relative to the noise. Trusting the BIM means correcting the pose with it, except for the wall or room that was attributed, whose change is only proposed for a person to check. Where the explanations cannot be separated, the BIM is left unchanged and anchoring is deferred in the ambiguous directions. Where no single fault explains the data, the BIM is left unused and the position is withdrawn.

\section{Related Work}
\label{sec:related}

A scan--BIM conflict sits at the intersection of four lines of work: localization against a floor plan or a BIM, localization that must tolerate discrepancies, Scan-vs-BIM comparison, and fault detection. Table~\ref{tab:related} places the closest methods against the decision made here.

\subsection{BIM as a localization map}
\label{sec:related-map}

Building models have been widely adopted for facility inspection and navigation~\cite{Chen2023FacilityInspection, Pauwels2023LiveSemanticDT}. For localization, the role is narrower. Boniardi et al.\ track a LiDAR through CAD floor plans over long periods and keep the map current with pose-graph mapping and Bayesian filtering~\cite{Boniardi2019CADLocalization}. Li et al.\ localize online on imprecise floor plans by scan matching~\cite{Li2020ImpreciseMaps}, and FP-Loc matches projected walls and pillars to the plan from a given initial pose, estimating gravity and sensor height from the ceiling~\cite{Gao2022FPLoc}. Yin et al.\ localize a 3D LiDAR on semantic maps generated from the BIM~\cite{Yin2023SemanticBIM}, and Hendrikx et al.\ connect BIM semantics to 2D LiDAR localization~\cite{Hendrikx2021SemanticBIM}. Situational graphs arrange walls and rooms in a hierarchy~\cite{Bavle2022SGraphs, Bavle2023SGraphsPlus}. This is the representation that iS-Graphs and diS-Graphs build on. Wang et al.\ fold BIM geometry into factor-graph LIO for construction robots~\cite{Wang2026iBIMLIO}. In every case, the model is the reference.

When the layout is symmetric, however, several matches can be equally plausible. Qiao et al.\ register a scan to the BIM by pose Hough voting, checked with an occupancy-aware score~\cite{Qiao2025SpeakSameLanguage}. iS-Graphs matches an architectural graph to the online situational graph and, in a symmetric environment, when the best cluster still holds several optimal matches, waits for more rooms and walls until the match is unique~\cite{Shaheer2023ISGraphs}. Zhang et al.\ check proposed LiDAR-to-BIM alignments against competing explanations before accepting one as the global initialization~\cite{Zhang2026MoveFirst}. All three methods decide where the robot is when tracking starts. This paper instead addresses a different question that arises during tracking: whether a scan--BIM conflict can be attributed to the odometry prior, a single wall, or a whole room, and whether the BIM may be revised from that attribution.

\subsection{Handling scan--BIM discrepancies in localization}
\label{sec:related-discrepancy}

At each conflict, the method in this paper tests whether the cause is attributable, and it proposes a BIM revision only when the cause can be uniquely determined. diS-Graphs~\cite{Shaheer2025DiSGraphs} estimates the global alignment from walls without deviation, then the wall and room deviations, precisely because a single joint optimization confuses pose error with structural error.

Whether any BIM structure is allowed into the estimate is the second decision. BIM-Loc~\cite{Zhang2026BIMLoc} addresses this with multi-hit ray casting for scan-to-BIM association and hierarchical Bayesian inference that propagates discrepancy evidence from surface pixels to whole structures. Each patch keeps an inconsistency posterior, a binary indicator lets only structure judged consistent into the pose residuals, and the estimator does not depend on a particular odometry front end. What BIM-Loc decides is whether a piece of structure is both present and consistent. Our method decides which explanation produced a discrepancy, and when the BIM can be trusted. 

LIO-BIM~\cite{Stuehrenberg2025LIOBIM} takes a related approach: it couples LIO to the BIM and adds a BIM factor only when scan-to-BIM matching has converged with sufficient quality. If the test fails, the estimate stays with LIO.

Changing the map, rather than gating it, is the third response. On the planned BIM components, Spinner and Degani run particle-filter SLAM, estimate each component's existence probability with an exact inverse sensor model, and fall back to Monte Carlo localization where the building is already known~\cite{Spinner2024OnlineBIM}. BIM-SLAM builds an initial session from the BIM, anchors later sessions to it, and reconstructs elements the BIM does not contain~\cite{VegaTorres2023BIMSLAM}. SLAM2REF anchors multi-session trajectories to a BIM or a reference point cloud and updates the map by positive and negative change detection, tolerating small scan-to-map deviations~\cite{VegaTorres2024SLAM2REF}. These methods track whether elements exist and whether they have changed. Our question is a different one: the geometric offset of a wall or room that is still in place, and the choice between that offset and an odometry error.

Before BIM localization, there was already the habit of down-weighting an unreliable constraint. Switchable constraints reduce or remove outlier edges in a pose graph and do not name the physical source of the outlier~\cite{Suenderhauf2012Switchable}. Zhang et al.\ solve an optimization-based estimate only in well-conditioned directions~\cite{Zhang2016Degeneracy}, and X-ICP constrains or skips the update in directions it finds degenerate~\cite{Tuna2024XICP}. In these methods, conditioning chooses those directions; here, a direction is left to odometry when the candidates disagree about its BIM-anchored position.

To our knowledge, none of the LiDAR--BIM methods reviewed here tests, for each scan--BIM conflict during tracking, whether the conflict is attributable to the odometry prior, a single wall or a whole room, and withholds the BIM revision when the cause is not attributable.

\subsection{Scan-vs-BIM deviation analysis}
\label{sec:related-scanvsbim}

Scan-vs-BIM comparison checks what was actually built. As-built dimensions for compliance control are derived from recognized CAD objects~\cite{Bosche2010DimensionalControl}, and the same idea locates mechanical, electrical and plumbing (MEP) components that were not built where they were designed~\cite{Bosche2015ScanBIM}. Registration is usually one stage of that pipeline~\cite{Bosche2010DimensionalControl, Bosche2012PlaneRegistration, Zhang2024GlobalBIMRegistration, Liu2024DepthInformedRegistration}, though MEP sleeves have also been checked without it~\cite{Alsuhaibani2026RegistrationFree}. Object-level matching follows design-intent instances through changes of position, orientation and scale~\cite{Hu2024DesignIntent}, and an object-aware pipeline can turn the design model into an as-built twin automatically~\cite{Wang2025InformativeAsBuilt}.

Where a deviation came from, and how sure the comparison is, occupy a smaller set of studies. In offline settings, Anil et al.\ read the source of error in an as-is BIM, whether from collection, processing or modeling, from the pattern of the deviations, together with the magnitude and the kind of error~\cite{Anil2013DeviationAnalysis}. Meyer et al.\ combine point accuracy, surface properties and BIM uncertainty in Dempster--Shafer theory to decide whether a photogrammetric cloud confirms the model at a stated accuracy~\cite{Meyer2023BIMVerification}. Ers\"{o}z and Bosch\'e score component-level matches with coverage, distribution and distance, accounting for occlusion~\cite{Ersoz2026MatchingConfidence}. Luo et al.\ use the uncertainty of a segmentation to assess quality without ground truth~\cite{Luo2025UncertaintyQuality}, and Vassilev et al.\ estimate that uncertainty with Bayesian deep learning~\cite{Vassilev2024UncertaintySegmentation}. For the online setting, the comparison is performed on a moving robot and inside one measurement model that has to separate an LIO bias, a wall-surface offset and a room translation.

Digital-twin construction treats the BIM as semantically incomplete while the building is going up~\cite{Boje2020SemanticDigitalTwin}, and reference architectures keep planned information, raw monitoring data and derived knowledge in separate stores~\cite{Schlenger2025DTCArchitecture, Kosse2025SemanticDigitalTwins}. A conflict and its candidates are monitoring data. Only a revision that has been checked by a person can enter the model.

\subsection{Fault detection, identification and integrity}
\label{sec:related-dia}

Faulty measurements in geodesy are handled in the detection–identification–adaptation (DIA) framework. From DIA, Teunissen gives the distribution of the resulting estimator~\cite{Teunissen2018DIA}, and when hypotheses cannot be separated, Zaminpardaz and Teunissen show that the whole parameter vector cannot be adapted, while some functions of the parameters still can~\cite{Zaminpardaz2019Identifiability}. Therefore, a pose under several candidates is corrected only along the directions on which they agree. The minimal detectable bias (MDB) describes whether a fault can be detected; with several hypotheses it does not say whether the fault can be identified, and the minimal identifiable bias (MIB) is computed by sampling~\cite{Zaminpardaz2023MDBMIB}. Detection and attribution stay apart, under those two names, in Section~\ref{sec:identifiability}.

Integrity monitoring carries the same fault detection into navigation. Solution-separation and residual-based receiver autonomous integrity monitoring (RAIM) are compared by Joerger et al.~\cite{Joerger2014RAIM}. Robotic integrity monitors have been built for Kalman localization with data-association faults~\cite{DuenasArana2020Integrity}, for fixed-lag smoothing~\cite{AbdulHafez2020FixedLag}, and for EKF-SLAM with measurement faults and nearest-neighbor association errors~\cite{AbdulHafez2025SLAMIntegrity}. Innovation projections describe the effect of a wrong association and reduce its risk~\cite{Joerger2025InnovationProjections}. Rife et al.\ review integrity for georeferenced LiDAR across three architectures~\cite{Rife2025LidarIntegrity}, and Antonante et al.\ identify perception faults at runtime with diagnostic graphs~\cite{Antonante2023Monitoring}. These monitors exclude the faulty measurement or declare the position unavailable. If a method may also decline to answer, the relevant comparison is the error--reject trade-off~\cite{Chow1970RejectTradeoff} and the risk--coverage trade-off of selective classification~\cite{ElYaniv2010Selective}, which is why Section~\ref{sec:design} compares wrong attributions at the same answer rate.

Relative to DIA and RAIM, the fault hypotheses here are the walls and rooms of the BIM and the membership of walls in rooms, so whether a cause can be told apart depends on which witnesses the robot has seen. When it cannot, the candidates are kept and BIM anchoring is deferred until a new witness appears.

\begin{table*}[t]
\centering
\caption{Comparison with the prior work on scan–BIM conflict attribution. \ding{55}: no; \ding{51}: yes.}
\label{tab:related}
\small
\begin{tabular}{@{}*{5}{>{\raggedright\arraybackslash}p{0.17\textwidth}}@{}}
\toprule
Work & What it estimates & Tests, per conflict, whether the cause is attributable & Output when the cause cannot be told apart & Role in this paper \\
\midrule
diS-Graphs~\cite{Shaheer2025DiSGraphs} & Global alignment and wall and room deviations & \ding{55}; staged estimation starting from walls without deviation & One estimate of pose and deviations & Main baseline (reproduced) \\
BIM-Loc~\cite{Zhang2026BIMLoc} & Per-patch structural consistency that gates BIM constraints & \ding{55}; per-patch gating, no offset size or attribution & Localization continues with the gated structure & Real-data baseline (reproduced) \\
LIO-BIM~\cite{Stuehrenberg2025LIOBIM} & Pose; BIM factor added only after a match passes thresholds & \ding{55}; match-quality thresholds & Accepts or rejects the match by its quality; one pose either way & Gate reproduced \\
Floor-plan and BIM localization~\cite{Boniardi2019CADLocalization,Li2020ImpreciseMaps,Gao2022FPLoc,Yin2023SemanticBIM,Hendrikx2021SemanticBIM,Shaheer2023ISGraphs} & Pose, with the plan or BIM as the map & \ding{55}; iS-Graphs waits for a unique match in global localization & One pose, with the model as the reference & Background \\
BIM update and change detection~\cite{Spinner2024OnlineBIM,VegaTorres2023BIMSLAM,VegaTorres2024SLAM2REF,Wang2025InformativeAsBuilt} & Component existence, elements missing from the model, object-level as-built update & \ding{55} & Updates or flags the model & Background \\
Scan-vs-BIM~\cite{Bosche2010DimensionalControl,Anil2013DeviationAnalysis,Meyer2023BIMVerification} & Component recognition, deviation and quality & \ding{55}; error sources identified offline from deviation patterns~\cite{Anil2013DeviationAnalysis} & -- & Background \\
DIA and integrity~\cite{Teunissen2018DIA,Zaminpardaz2019Identifiability,Joerger2014RAIM,Rife2025LidarIntegrity} & Fault detection, identification and reliability & Theory of identifiability, MDB and MIB & Excludes the fault or declares the position unavailable & Theoretical basis \\
This work & Attribution among LIO bias, wall-surface offset and room translation & \ding{51} & Keeps the candidates, proposes no BIM revision and defers BIM anchoring in the ambiguous directions & -- \\
\bottomrule
\end{tabular}
\end{table*}

\section{Problem Formulation}
\label{sec:problem}

To begin with, we define the quantities the method works with. The state of each keyframe in a window is its planar position and yaw; height, roll, and pitch stay with the LIO. The three candidate explanations are then written as biases on a single residual, and we state the three reports the method must produce for every window, which will be detailed later.

\subsection{State, motion and window prior}
\label{sec:problem-state}

The method works on a path window of $N$ keyframes, with state
\begin{equation}
X=(x_0^\top,\ldots,x_{N-1}^\top)^\top,\qquad x_f=(t_f^\top,\theta_f)^\top,
\label{eq:state}
\end{equation}
where $t_f\in\mathbb{R}^2$ and $\theta_f$ are the position and yaw of keyframe $f$ in the BIM frame, respectively. LIO supplies the relative motion $z_{ij}$ between consecutive keyframes $i$ and $j$, expressed in the frame of keyframe $i$ and predicted by
\begin{equation}
h_{ij}(X)=\begin{bmatrix}R(\theta_i)^\top(t_j-t_i)\\ \operatorname{wrap}(\theta_j-\theta_i)\end{bmatrix},
\label{eq:odometry}
\end{equation}
where $R(\theta)$ is the planar rotation matrix and $\operatorname{wrap}$ maps an angle into $[-\pi,\pi)$. Here, the associated covariance is denoted by $\Sigma_{ij}$.

In addition to the relative motion, absolute information enters through a prior on the first keyframe, given by
\begin{equation}
e_P(X)=\begin{bmatrix}t_0-z_{P,t}\\ \operatorname{wrap}(\theta_0-z_{P,\theta})\end{bmatrix},
\label{eq:prior}
\end{equation}
here, the associated covariance is denoted by $\Sigma_P$. The prior defines the datum, that is, the coordinates of the start. On SLABIM, these coordinates come from a BIM alignment of the opening keyframes; in the laboratory they come from a laser tracker. The datum is passed from window to window by centering the next prior on the last working pose, the pose the method is about to adopt, so that it is carried forward by the odometry and grows softer with distance. In simulation, the prior is given directly.

\subsection{Wall measurements}
\label{sec:problem-walls}

In the BIM, each wall $w$ is represented as a line with unit normal $n_w$ passing through a point $p_w$. For a scan point $q_{fk}$ of keyframe $f$ associated with wall $w$, the signed point-to-line residual that measures the deviation of the point from the BIM wall is given by
\begin{equation}
\rho_{fwk}(X)=n_w^\top\big(R(\theta_f)\,q_{fk}+t_f-p_w\big),
\label{eq:wall-residual}
\end{equation}
which vanishes when the point lies on the BIM wall.

This residual is subject to noise from two sources.
Projected onto the wall normal, range noise of standard deviation $\sigma_r$ along the beam contributes the variance $v_{fwk}$ (\ref{app:frontend}), and $\sigma_t$, the ordinary construction tolerance of that wall, is shared by every point of the wall inside the window.
The covariance of the residuals of wall $w$ is
\begin{equation}
\Sigma_w=\operatorname{diag}(v_{fwk})+\sigma_t^2\,\mathbf{1}\mathbf{1}^\top.
\label{eq:wall-noise}
\end{equation}

The prior, the odometry, and the measurements of different walls are independent of one another. Whitening each block and stacking them produces the residual $r(X)$ and the cost $J(X)=\lVert r(X)\rVert^2$. Numerical values are given in \ref{app:lifetest} and~\ref{app:frontend}.

\subsection{Fault classes}
\label{sec:problem-faults}
\begin{figure}[t]
\centering
\includegraphics[width=\linewidth]{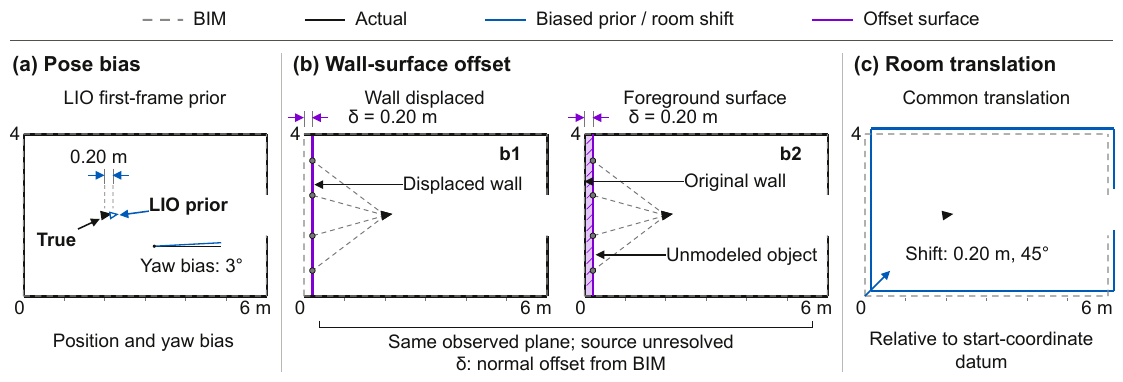}
\caption{The three fault classes of the single-fault model, each shown at 0.20~m. (a) LIO bias: the LIO pose is offset in $x$, $y$ and yaw (three parameters). (b) Wall-surface offset: the surface observed at one BIM wall is offset along the wall normal by $\delta$ (one parameter). It may come from an as-built deviation of the wall (b1) or from an unmodeled object in front of it (b2); the wall plane alone cannot separate the two, and the method does not distinguish them. (c) Room translation: the whole room is translated in the plane (two parameters) relative to the datum, the start coordinates propagated through the window priors. Normal construction tolerance is treated as noise, not as a fault.}
\label{fig:fault-classes}
\end{figure}

With the residual model of Sections 3.1 and 3.2 in place, a conflict can be explained by one of three fault classes. These classes are illustrated in Fig.~\ref{fig:fault-classes}. Explanation $h$ adds a bias $b_h$ through a fixed design matrix $C_h$, so that the residual under explanation $h$ is
\begin{equation}
r_h(X,b_h)=r(X)+C_h b_h,
\label{eq:fault-model}
\end{equation}
The null explanation H0 corresponds to the case with no fault, that is, $b_h=0$. The three fault classes are defined as follows.

An \emph{LIO bias} $b_L=(b_x,b_y,b_\theta)^\top$ is an error in the absolute pose that LIO brings into the window. Acting on the prior of the first keyframe, it turns the whitened residual into $W_P(e_P+b_L)$, where $e_P$ is the prior residual and $W_P$ is the whitening factor of $\Sigma_P$, while the odometry and the scans stay unchanged. Without an absolute prior there is no LIO-bias explanation. On the bias itself there is no magnitude bound.

A \emph{wall-surface offset} $\delta_w$ moves the surface observed at wall $w$ along $n_w$, so $C_w$ is nonzero only in the rows of that wall. In this paper, we assume the offset is bounded. The surface may be the wall itself, built away from the design, or an object the BIM does not contain, and the wall plane gives the same observation either way.

A \emph{room translation} $b_R=(\Delta x,\Delta y)^\top$ shifts every wall of room $R$ relative to the datum. The design matrix is defined as $C_R=\sum_{w\in\mathcal{W}_R}c_w\,n_w^\top$, where $\mathcal{W}_R$ is the set of walls the BIM assigns to the room and $c_w$ is the column of a unit offset of wall $w$. The translation has no magnitude bound.

Under the single-fault assumption, only one fault is active at a time and the explanations are fitted separately, so that measurements requiring two faults at once fit none of them; a wall marked as a fixed reference has no wall explanation of its own and does not move with its room. The hypothesis set $\mathcal{H}$ contains one wall-surface offset for each visible wall that is not fixed, one room translation for each visible room and, when the window has an absolute prior, one LIO bias, with size $m=\lvert\mathcal{H}\rvert$ excluding H0.

\subsection{Required outputs}
\label{sec:problem-outputs}

Each window returns the three reports separately, because the answers can overlap. Under H0, detection says whether the scan conflicts with the BIM. The cause of a detected conflict is a unique attribution when one explanation remains, an ambiguous cause together with its candidates when several remain, or all explanations excluded when none remains. Availability, on each coordinate of the last keyframe, is either the working pose or a mark that the coordinate is unavailable, and a direction the BIM cannot verify is marked unverified when there is no conflict. Beside the pose comes the union interval, the union of the pose intervals of every compatible explanation. Only a unique wall or room attribution produces a BIM-side output, and that output is a proposed revision.

\section{When Can Pose Errors and As-Built Deviations Be Told Apart?}
\label{sec:identifiability}

Section~\ref{sec:problem} places the three explanations onto a single residual. This section concentrates on when those explanations can be told apart. In the single-room setting of Fig.~\ref{fig:paired-worlds}, a pure translation can be produced either by a room translation or by an LIO bias, while the LIO bias can additionally produce a yaw that a room translation cannot. Whether two explanations can be distinguished in principle is a geometric property and is unchanged by collecting more data under the same geometry; whether a particular stretch of trajectory separates them also depends on the size of the fault relative to the noise. The argument developed here applies the detection--identification--adaptation (DIA) identifiability framework~\cite{Teunissen2018DIA, Zaminpardaz2019Identifiability} to these fault classes.

\subsection{Projected fault subspaces}
\label{sec:identifiability-subspaces}

At a pose $\bar X$, linearize: residual $\bar r=r(\bar X)$, Jacobian $A=\partial r/\partial X$. Inside the range of $A$ is where a change of pose moves the residual, so any part of a fault signature that lies in that range can be absorbed by adjusting the pose. What remains visible is the projection onto the orthogonal complement, which can be expressed as
\begin{equation}
P^\perp=I-AA^{+},\qquad S_h=\operatorname{range}(P^\perp C_h),
\label{eq:projected-subspace}
\end{equation}
where $A^{+}$ is the pseudoinverse and $S_h$ is the projected fault subspace of explanation $h$. A fault $h$ is detectable when $S_h\neq\{0\}$, because some of its signature then survives every pose adjustment. Walls and references that constrain the pose along the offset direction, counting the offset wall itself, are the witnesses. For a wall-surface offset, $S_h$ is nonzero only when the offset direction has at least one witness besides that wall.

In addition, the magnitude bound restricts the set of signatures that an explanation can produce, which can be expressed as
\begin{equation}
E_h=\{P^\perp C_h b:\ b\in\mathcal{B}_h\},
\label{eq:explainable-set}
\end{equation}
where $\mathcal{B}_h$ is a bounded interval for a wall-surface offset and the full parameter space for a room translation or an LIO bias. For a room or the LIO, $E_h$ coincides with the subspace $S_h$; for a wall, it is a segment of the line $S_h$. The least-squares cost of explanation $h$ is the squared distance from the projected residual to this set, namely,
\begin{equation}
J_h=\min_{\Delta X,\,b\in\mathcal{B}_h}\lVert\bar r+A\Delta X+C_hb\rVert^2=\operatorname{dist}\big(P^\perp\bar r,\,E_h\big)^2.
\label{eq:cost-distance}
\end{equation}
In the noise-free case, the explanation accounts for the data when this distance is zero. Once noise is present, Section~\ref{sec:method} replaces the zero-distance condition with a statistical test fixed in advance of the data.

\subsection{Detection versus identification}
\label{sec:identifiability-identification}

Detection against the null hypothesis H0 requires only a nonzero $S_h$, whereas identification against one another requires more. For a fault to be attributable, its signature must lie outside the explainable set of every other explanation, so the test examines every candidate rather than a single pair. A fault may be detectable yet still not attributable. The following four relations then arise.
\begin{itemize}
\item \emph{Coincidence}: $E_h=E_j$. Every observation fits them equally, and no method can tell them apart.
\item \emph{Nesting}: $E_h\subseteq E_j$. Whatever the narrower explanation $h$ fits, the broader explanation $j$ fits as well (Section~\ref{sec:identifiability-nesting}).
\item \emph{Partial overlap}: $S_h\cap S_j\neq\{0\}$ without nesting. The two can be separated or not according to whether the present signature lies in the overlap.
\item \emph{Separation}: $S_h\cap S_j=\{0\}$. The two can be told apart on every signature.
\end{itemize}
In a single room the room translation is nested in the LIO bias: a pure translation lies in the intersection of the two explainable sets, while a yaw is explained only by the LIO bias. When several explanations remain, the pose cannot be corrected as a whole, because they disagree about it. As noted in Section~\ref{sec:related-dia}, only some functions of the parameters can then be adapted~\cite{Zaminpardaz2019Identifiability}; for the pose, these are the directions in which every retained explanation yields the same BIM-anchored position. This leads to the rule of deferred BIM anchoring.

\subsection{Asymmetric nesting}
\label{sec:identifiability-nesting}

The inclusion is asymmetric, as illustrated in Fig.~\ref{fig:witness-nesting} (a): a room translation can never be confirmed from that room alone.
\begin{proposition}[Asymmetric nesting]
\label{prop:nesting}
Let $h$ and $j$ be two explanations with $E_h\subseteq E_j$ in the linearized model. Then (a) $J_j\le J_h$ for every residual $\bar r$, so every observation that $h$ explains is also explained by $j$; (b) in the noise-free case, if $P^\perp\bar r\in E_j\setminus E_h$, then $j$ explains the observation and $h$ does not.
\end{proposition}
Proof: Please refer to \ref{app:proof}.
A narrower explanation can therefore never be confirmed on its own: whenever it fits, the broader one fits as well, and the conflict remains ambiguous. When the data contain a component that the narrower explanation cannot produce, the narrower one is excluded and the broader one remains. Only after every other candidate has also been excluded is the broader explanation reported. Because $E_h$ includes the magnitude bound, nesting is a relation between explainable sets, not between subspaces alone. A wall whose projected subspace lies inside that of its room is nested in the room. The room is not nested in the wall, since a sufficiently large room offset along the wall’s normal is explained by the room translation and not by the wall-surface offset.

\subsection{Witness ladder}
\label{sec:identifiability-ladder}

\begin{figure}[t]
\centering
\includegraphics[width=\linewidth]{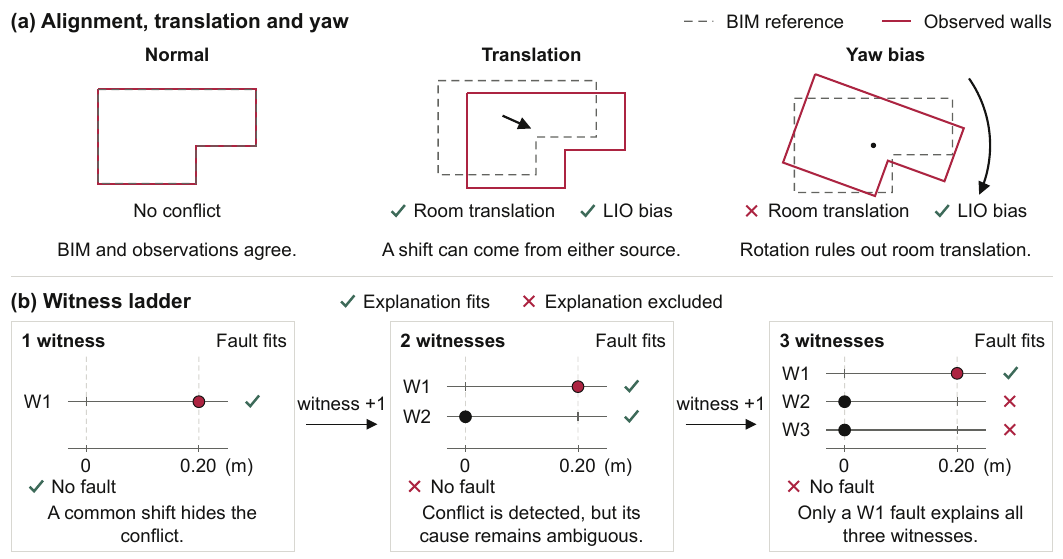}
\caption{(a) Single-room view: an aligned scan raises no conflict; a translation is explained by a room translation and by an LIO bias alike; a rotation is explained only by the LIO bias, which adds yaw. Anything a room translation explains, an LIO bias explains as well, so a room translation can never be confirmed on its own. (b) Witness ladder, Manhattan special case. Each line is one witness along the offset direction, and its dot marks the robot position that witness implies. With one witness, a fault produces no contradiction; with two, it is detected but cannot be attributed; with three or more, it can be attributed.}
\label{fig:witness-nesting}
\end{figure}

Manhattan layouts reduce the subspaces to a simple count of witnesses, as illustrated in Fig.~\ref{fig:witness-nesting} (b). Each wall whose normal lies along the offset counts as one witness, and a prior that constrains the pose along the offset counts as one more, with strength set by its standard deviation. With one witness, the pose absorbs the offset and no conflict appears. With two witnesses, the conflict is detected, but an offset of the first wall produces the same data as an offset of the second wall together with a pose shift, so the cause cannot be attributed. With three or more witnesses, the walls that agree single out the one that does not. The count is a special case for orthogonal walls; non-orthogonal walls and the strength of the prior remain subject to the projected-subspace test.

Rooms follow the same counting principle, because a room translation is defined relative to the datum. When two rooms disagree inside one window, a conflict is detected. The explanations ``the first room moved'' and ``the second room moved'' differ only through the prior, so an offset that is small relative to the prior remains ambiguous and BIM anchoring is deferred. A third group of walls that agrees with one of the two rooms makes the cause attributable; this group may be another room, another region, or walls that belong to no room. Fixed references, such as the walls outside the door in the simulated building, are sufficient once seen, whereas real sequences contain no fixed wall, so the third group must consist of walls that the robot actually observes.

\subsection{Magnitude relative to noise}
\label{sec:identifiability-noise}

Noise adds a second requirement: the fault must be large enough relative to the noise in the window at hand. The minimum detectable bias (MDB), the smallest fault detected with a stated probability, and the minimum identifiable bias (MIB), the smallest fault attributed correctly with a stated probability~\cite{Zaminpardaz2023MDBMIB}, both depend on direction when the fault has several parameters. The method decides that an explanation fails using the test of Section~\ref{sec:method}, so a fault large enough to be detected can still be too small to be attributed, and the cause then remains ambiguous. 

A new reference ends an ambiguity when it constrains the ambiguous direction; new rows in $A$ then change the projected subspaces, and explanations that previously coincided or nested can become separable. The method in Section~\ref{sec:method} waits for such a recovery before making an attribution.

\section{Identifiability-Aware LiDAR-Inertial--BIM Fusion}
\label{sec:method}

\begin{figure*}[t]
\centering
\includegraphics[width=0.8\linewidth]{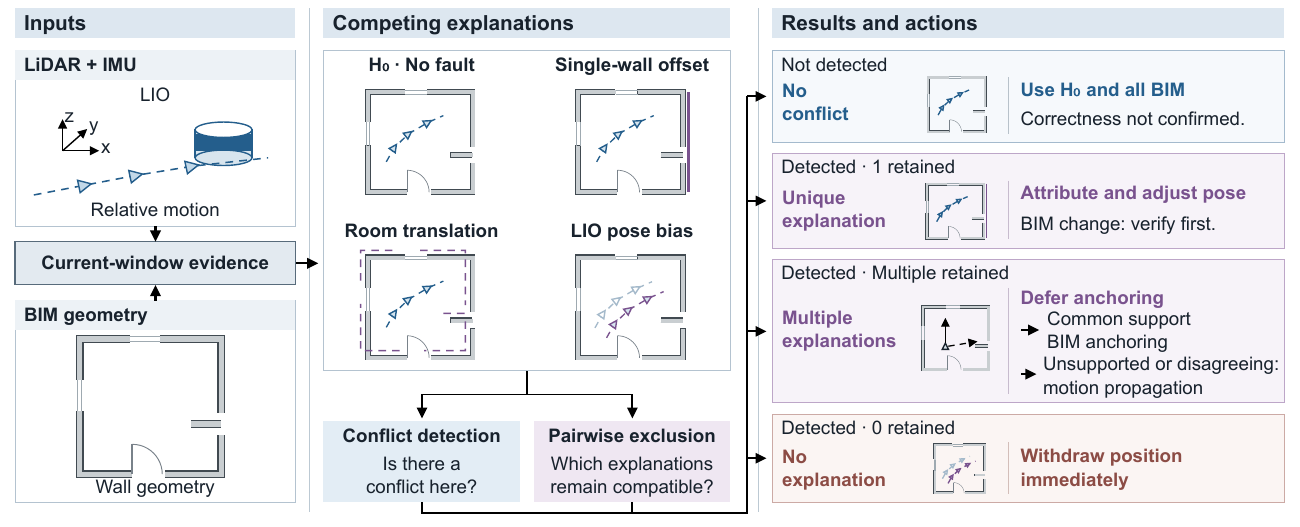}
\caption{Pipeline of the identifiability-aware LiDAR-inertial--BIM fusion. LiDAR-inertial odometry runs without the BIM; the BIM enters through wall-patch association in fixed-length path windows. All single-fault explanations (H0, each visible wall, each visible room and an LIO bias) are fitted, a conflict is detected against H0, and explanations are excluded pairwise. If one explanation remains, it is attributed and its BIM change is only proposed for verification. If several remain, they are retained and BIM anchoring is deferred in the directions where they do not agree on a BIM-anchored position. If none remains, the position is withdrawn. The pose interval is the union over compatible explanations, an ambiguous axis is withdrawn after a preset distance, and a new reference triggers re-identification.}
\label{fig:pipeline}
\end{figure*}

Every window answers the three questions posed in Section~\ref{sec:problem-outputs}, and the answers may overlap: whether the scan conflicts with the BIM, what caused a conflict, and whether each coordinate is available. A window with no conflict can still mark a direction as unverified. The overall pipeline is shown in Fig.~\ref{fig:pipeline}, and the order of operations inside a window is given by Algorithm~\ref{alg:window}. Parameter values are listed in Tables~\ref{tab:params-sim} and~\ref{tab:params-real}. The text below states the rules.

\subsection{Pipeline}
\label{sec:method-pipeline}

LiDAR-inertial odometry runs independently. Only the wall measurements of Section~\ref{sec:problem-walls} admit the BIM into the estimation, leaving the odometry as an independent witness. The datum is fixed at the start of a run and then carried forward. On SLABIM, the opening keyframes are aligned to the BIM once and are not tested; that alignment serves as the datum, and the start region is taken to agree with the BIM. The room in which the run starts, however, remains an ordinary room in $\mathcal{H}$. Treating it as a fixed reference would turn an offset of the start room into an apparent offset of every later room. In the laboratory, a laser-tracker pose serves as the datum, and testing begins at the first keyframe. Association thresholds for the alignment are given in \ref{app:frontend}.

Keyframes are spaced at regular intervals and grouped into fixed-length windows, so that walls observed at different times are linked by the relative motion. Consecutive windows share a boundary keyframe, and an output is produced after every keyframe using the portion of the window received so far. Scan points are associated at the pose predicted from the previous working pose and the LIO motion. On the real sequences, the points of a wall are grouped into patches, and a wall enters the window when most of its patches agree. Furniture, people, and objects that cover only part of a wall are excluded from the measurements, whereas an object that covers most of a wall appears as a wall-surface offset of that wall. The patch rules are given in \ref{app:frontend}.

Every explanation in $\{\mathrm{H0}\} \cup \mathcal{H}$, the null hypothesis together with all single-fault explanations, is then fitted independently by Gauss--Newton, starting from the same predicted pose and from zero bias; each retains its own pose, bias, and covariance. A joint fit would constitute a multiple-fault estimate, which the model of Section~\ref{sec:problem-faults} does not support. A wall-surface offset whose unconstrained value exceeds the magnitude bound is held at the bound while the pose is refitted.

\subsection{Detection and pairwise exclusion}
\label{sec:method-tests}

The tests use the subspaces of Section~\ref{sec:identifiability} evaluated at the fitted poses. Detection employs the Jacobian $A_0$ at the fit of H0, while pairwise exclusion employs each candidate’s Jacobian at its unconstrained fit. Detection checks whether any explanation reduces the cost of H0 by more than would be expected from noise alone. For explanation $h$, the reduction $D_h=\max(0,J_0-J_h)$ is compared with a $\chi^2$ distribution whose degrees of freedom $q_h=\operatorname{rank}\big((I-A_0A_0^{+})C_h\big)$ equal the number of directions of the fault that the pose cannot absorb, which yields
\begin{equation}
p_h=\Pr\big(\chi^2_{q_h}\ge D_h\big),\qquad
\text{conflict if}\ \min_h p_h<\frac{\alpha}{\max(1,m_+)},
\label{eq:detection}
\end{equation}
where $m_+$ is the number of explanations with $q_h>0$, and $\alpha$ is a preset significance level. An explanation with $q_h=0$ cannot raise the alarm, yet it remains a candidate in the exclusion tests; this is how a room translation inside a single room stays available to compete with the LIO bias.

Exclusion checks, for every candidate, whether some competitor explains a direction that the candidate cannot. Let $B_i$ be an orthonormal basis of the range of candidate $i$’s Jacobian at its unconstrained fit, including its bias columns, and let $U_{j|i}$ be an orthonormal basis of the directions that competitor $j$ adds. Candidate $i$ is excluded by $j$ when
\begin{equation}
T_{i,j}=\lVert U_{j|i}^\top r_i^{\mathrm{free}}\rVert^2+L_i>\chi^2_{\nu_{i,j},\,1-\alpha/m^2},
\label{eq:exclusion}
\end{equation}
where $\nu_{i,j}$ is the dimension of $U_{j|i}$ (plus one if $i$ is a wall), $r_i^{\mathrm{free}}$ is the residual of the unconstrained fit, and $L_i=\max(0,J_i-J_i^{\mathrm{free}})$ is the cost of holding a wall offset at its magnitude bound. Nesting, as established in Section~\ref{sec:identifiability-nesting}, appears in the first term: a competitor that adds no direction cannot exclude the candidate, so a broader explanation remains when a narrower one fits. The magnitude bound appears in the second term, and a wall can be excluded by its bound loss alone, even in the case $m=1$ where it has no competitor. The threshold that covers every ordered pair is $\alpha/m^2$. No fault class is preferred.

After the tests, a single remaining explanation constitutes a unique attribution, several remaining explanations constitute an ambiguous cause, and none remaining means that all explanations have been excluded. The output is “no conflict” when detection did not fire. The loop over pairs is given in Algorithm~\ref{alg:window}.

\subsection{Deferred BIM anchoring}
\label{sec:method-anchoring}

One room and a door under a strong prior illustrate the rule. Before the wall outside the door becomes visible, the robot sees only one room, so a translation fits both a room translation and an LIO bias, and both candidates remain. Along the offset they do not agree on a BIM-anchored position, and that direction follows odometry. Directions still determined by the walls keep the BIM. Once the outside wall comes into view it acts as a third witness, the subspaces separate, and the pose is corrected. Without such a wall, the ambiguous direction is withdrawn after the distance limit described below. Under a weak prior the same room translation goes undetected, so there is no conflict and the pose follows the BIM; the affected coordinates are marked unverified. Table~\ref{tab:update} summarizes this behavior for every outcome, because a pose is still required on every keyframe, including while the cause remains ambiguous.

No conflict uses the fit of H0 and every wall, and a direction that a compatible explanation cannot anchor with the walls and the relative motion is marked unverified. A unique attribution corrects the pose according to that explanation; a wall or room attribution becomes a proposed BIM revision when the estimated offset exceeds a small threshold, and an LIO attribution is reported as an error of the robot. Several candidates split the directions of the pose into those they agree on and those they do not. A direction is ambiguous when at least one candidate cannot anchor it with the walls and the relative motion, or when the candidates anchor it at positions that differ by more than their uncertainties. For candidates $i$ and $j$ with fitted poses $\hat X_i$ and $\hat X_j$, the second test is applied along their difference $d_{ij}$, after removing the directions of the first test,
\begin{equation}
\lVert d_{ij}\rVert>z_{0.975}\,(\sigma_i+\sigma_j),
\label{eq:disagreement}
\end{equation}
where $\sigma_i$ and $\sigma_j$ are the standard deviations of the two fits along that direction. The sum bounds the standard deviation of the difference when the cross-covariance is unknown. With $U_{\mathrm{work}}$ an orthonormal basis of the ambiguous directions, $P_{\mathrm{work}}=I-U_{\mathrm{work}}U_{\mathrm{work}}^\top$ and $X^{\mathrm{pred}}$ the pose predicted from the prior and the odometry, the working pose is
\begin{equation}
X^{\mathrm{work}}=X^{\mathrm{pred}}+P_{\mathrm{work}}\big(\hat X_c-X^{\mathrm{pred}}\big),
\label{eq:working-pose}
\end{equation}
where $c$ is the remaining explanation of lowest cost. Along an ambiguous direction the position follows the prior and the odometry, so a BIM correction there waits until the cause is attributed. Along the other directions the candidates agree, so the lowest-cost explanation supplies the correction and the robot can keep moving, while the cause remains unsettled. A comparable split appears in GNSS/INS integration, where the inertial solution carries the position while a satellite measurement is suspect.

A coordinate becomes unavailable in any of three separate situations. When every explanation has been excluded, as when two discrepancies are present at once, every coordinate is reported unavailable, the interval is empty, and nothing is proposed for the BIM; the prior and the odometry still propagate a pose internally as a placeholder for the next window. A coordinate is also unavailable when it is anchored neither by the BIM, along a direction common to the remaining explanations, nor by the prior and the odometry. Passing the distance limit is the third situation.

The reported interval is the union of the intervals of the remaining explanations, so that every candidate is covered. For a projection $a^\top X$, the interval of one explanation is centered on that explanation's fit, with half-width $z_{0.975}\sqrt{a^\top V a}$ from its weighted least-squares covariance $V$, and it is unbounded when the explanation cannot determine $a^\top X$. The nominal level is 95\% along each projection. With no conflict, H0 joins the union, and its half-width is widened by the largest pose shift a single wall offset inside the magnitude bound could cause, so the interval also covers a wall offset too small to detect.

Because odometry error grows with distance, a counter accumulates the distance travelled while a coordinate stays ambiguous; beyond a preset distance under the simulation noise model that coordinate is reported unavailable. The candidates and their intervals remain, so a later reference wall can still resolve the ambiguity. The boundary cases of the counter are given in Algorithm~\ref{alg:window}.

\begin{table}[t]
\centering
\caption{Working pose and BIM output for each outcome of the tests.}
\label{tab:update}
\small
\begin{tabular}{@{}>{\raggedright\arraybackslash}p{0.2\linewidth}>{\raggedright\arraybackslash}p{0.45\linewidth}>{\raggedright\arraybackslash}p{0.25\linewidth}@{}}
\toprule
Outcome & Working pose & BIM \\
\midrule
No conflict & Fit of H0 with all walls; directions that the BIM cannot verify are marked as unverified & Unchanged \\
Unique attribution & Corrected according to the attributed explanation & Corrected copy proposed for verification; the original is not replaced \\
Ambiguous & Correction of the lowest-cost remaining explanation in the directions where all candidates agree; prior and odometry in the ambiguous directions & Unchanged \\
All explanations excluded & Position withdrawn and reported as unavailable & Unchanged \\
\bottomrule
\end{tabular}
\end{table}

\subsection{Recovery}
\label{sec:method-recovery}

Each new keyframe is followed by a fresh fit and a fresh test on the portion of the window received so far. Once a new reference constrains an ambiguous direction, the projected subspaces change, and explanations that could not previously be separated can become separable (Section~\ref{sec:identifiability-noise}). The poses and offsets of that window are then corrected with the attributed explanation. A cause does not carry over to the next window. What does carry over is the last working pose, which becomes the center of the next prior and thereby propagates the datum.

\subsection{Computational cost}
\label{sec:method-cost}

Each output requires $m+1$ fits, one per explanation including H0, and at most $m(m-1)$ pairwise tests. The count $m$ grows with the number of walls and rooms in view. Each fit is a least-squares problem in the $3N$ pose parameters and at most three bias parameters. Although the shared tolerance $\sigma_t$ makes the covariance of a wall's residuals dense, the weighted cost remains available from per-wall, per-keyframe sums of the point coordinates and of their products. A fit does not grow with the number of scan points. Runtime is reported in Section~\ref{sec:results-runtime}.

\begin{algorithm}[t]
\caption{Processing of one window after each new keyframe}
\label{alg:window}
\small
\begin{algorithmic}[1]
\Require window prior $(z_P,\Sigma_P)$ if any, odometry $\{z_{ij}\}$, associated wall points, walls and rooms of the BIM
\State $\mathcal{H}\gets$ one wall per visible wall that is not fixed, one room per visible room, LIO bias if a prior exists
\ForAll{$h\in\{\mathrm{H0}\}\cup\mathcal{H}$}
  \State fit $(\hat X_h,\hat b_h)$ from $X^{\mathrm{pred}}$; hold a wall offset at $\pm0.35$~m if it exceeds the bound
\EndFor
\State detected $\gets \min_h p_h<\alpha/\max(1,m_+)$ \Comment{Eq.~\eqref{eq:detection}}
\State $\mathcal{R}\gets\mathcal{H}$
\ForAll{ordered pairs $(i,j)$, $i\ne j$}
  \If{$T_{i,j}>\chi^2_{\nu_{i,j},1-\alpha/m^2}$} $\mathcal{R}\gets\mathcal{R}\setminus\{i\}$ \Comment{Eq.~\eqref{eq:exclusion}}
  \EndIf
\EndFor
\If{$\mathcal{H}=\{w\}$ for a wall $w$ and its bound loss is significant} $\mathcal{R}\gets\emptyset$
\EndIf
\If{not detected}
  \State no conflict: $X^{\mathrm{work}}\gets\hat X_0$; mark unverified directions; interval over $\mathcal{R}\cup\{\mathrm{H0}\}$
\ElsIf{$\lvert\mathcal{R}\rvert=1$}
  \State unique attribution: correct the pose; propose a BIM revision if the offset exceeds 0.05~m, or report the LIO bias
\ElsIf{$\lvert\mathcal{R}\rvert>1$}
  \State ambiguous: find the ambiguous directions; $X^{\mathrm{work}}$ by Eq.~\eqref{eq:working-pose}; interval over $\mathcal{R}$
\Else
  \State all explanations excluded: withdraw the position; empty interval
\EndIf
\State update the distance clocks of $x$, $y$ and yaw; report coordinates past the cap as unavailable
\Statex clear on attribution, or when the axis leaves the ambiguous directions; a short no-conflict gap does not clear the clock; excluding every explanation does not start one; on real sequences the clock restarts each window
\end{algorithmic}
\end{algorithm}

\section{Experimental Design}
\label{sec:design}

The experiments address five questions in order. When the cause cannot be told apart, does the method make fewer wrong attributions (RQ1)? When it can, does the method still attribute it (RQ2)? Does a new reference restore the attribution (RQ3)? Are the uncertainty statements honest (RQ4)? On real buildings, does the behavior follow Section~\ref{sec:identifiability} (RQ5)? The definitions used to answer them are fixed in this section.

\subsection{Data}
\label{sec:design-data}

The two-dimensional generator renders LiDAR scans with occlusion in four layouts: a corridor, a corner, a room with a door, and the same room with a pillar, as shown in Fig.~\ref{fig:sim-layouts} and detailed in \ref{app:lifetest}. Pillars and the two walls outside the door are fixed references. Faults are injected one at a time, with the ground truth known. Separate from the pre-registered study, a demonstration in the same room with a door follows deferred BIM anchoring through the paired worlds in five configurations, with 200 trials per world. One uses a weak prior with the reference wall at 8~m. The others use a strong prior, with the reference wall at 4, 8 or 16~m or with no reference wall. The specification also fixes a recovery grid and a paired-world grid, reported in Table~\ref{tab:recovery} and not part of the pre-registered criterion. A further 200 trials test recovery inside one fixed window, and Table~\ref{tab:data} lists what each source is used for.

Room 209 is a laboratory room with an IFC model, and its sequences were recorded. Sequence 209\_1 calibrates the noise and the transform between the laser tracker and the IFC. From a tracker pose, the test sequences 209\_2 and 209\_3 are processed from their first keyframe, and the tracker provides no per-frame ground truth. SLABIM~\cite{Huang2025SLABIM} provides LiDAR-inertial sessions and a mesh BIM of a university building. A per-frame reference is recovered by registering each raw scan to the released map point cloud, without the BIM (\ref{app:frontend}). The references of the two sessions on floor 1F differ by a rigid offset of about 18~cm, and the absolute accuracy is unknown. The frozen front end, which supplies initialization, associations and window priors, was developed on floor 1F and then frozen. Floors 3F and 4F were used only after that freeze, under specifications fixed before the runs. Results on 1F are development-floor results.

\begin{table*}[t]
\centering
\caption{Data, datum, truth and use of each data source.}
\label{tab:data}
\small
\begin{tabular}{@{}*{5}{>{\raggedright\arraybackslash}p{0.175\textwidth}}@{}}
\toprule
Data & BIM & Datum & Truth or reference & Use \\
\midrule
Two-dimensional generator & Corridor, corner, room, room with pillar; fixed references given & Given prior (strong, weak or none) & Known truth & Pre-registered study (comparison, recovery and paired-world grids); deferred-anchoring demonstration; fixed-window recovery test (RQ1--RQ4) \\
Laboratory room 209 (209\_1, 209\_2, 209\_3) & IFC model of the room & Laser-tracker initial pose for the test sequences 209\_2 and 209\_3; 209\_1 calibrates the noise and the tracker to the IFC & Initial pose only & Paired injections (RQ5) \\
SLABIM 1F (sessions 1F2, 1F3) & Mesh BIM & BIM alignment of the first nine keyframes & Per-frame reference; about 18~cm disagreement between sessions & Development floor: full sessions on the original BIM, paired injections (RQ4, RQ5) \\
SLABIM 3F and 4F & Mesh BIM & BIM alignment of the first nine keyframes & Per-frame reference recovered from the map & Paired injections, accessed after the front-end freeze (RQ4, RQ5) \\
\bottomrule
\end{tabular}
\end{table*}

\subsection{Baselines and ablations}
\label{sec:design-baselines}

The baselines for diS-Graphs and BIM-Loc are reproductions based on the original papers~\cite{Shaheer2025DiSGraphs,Zhang2026BIMLoc}, named diS-Graphs-style and BIM-Loc-style. Please refer to \ref{app:baselines} for details. Uncertainty-based abstention declines to answer when the largest standard deviation among the walls and rooms flagged by the diS-Graphs-style baseline exceeds a threshold. Sweeping this threshold produces the answer-rate curve.

For LIO-BIM, the official implementation requires orientation from the IMU and a BIM in IFC format. SLABIM provides a six-axis IMU and a mesh BIM, and the laboratory IMU is also six-axis, so the official implementation was not run on either dataset. The LIO-BIM-style gate reproduces the rule of using the BIM only after a good match. Please refer to \ref{app:baselines} for the six differences from the official system.

Ablations isolate parts of our method. The forced variant selects the lowest-cost candidate whenever several remain, and is otherwise identical. The trust-BIM ablation, labeled Trust BIM in Table~\ref{tab:injections} and the figures, treats every wall as correct and allows no fault, while keeping the LIO prior and the odometry. For a fair comparison, every method, including the baselines and the ablations, sees only the portion of the window that has been received so far.

\subsection{Labels, injections and metrics}
\label{sec:design-metrics}

This subsection defines the predicted labels, the fault injections, and the metrics used to evaluate the methods. Each condition carries a predicted label, computed before any method runs. In simulation the label is obtained at the true pose; on the real sequences it is fixed in advance of the unmodified runs. Under the declared linear noise model, a fault is labeled undetectable when its detection power at the level of Eq.~\eqref{eq:detection} is below 0.8, attributable when its power of separation from every competitor at the level of Eq.~\eqref{eq:exclusion} is at least 0.8, and detectable but not attributable otherwise. These labels are pairwise powers rather than the joint decision of the full test, so a fault labeled detectable but not attributable can still be attributed on a given run. A conflict is permanently unattributable only when the offset direction lies in the intersection of the explainable sets.

In the two-dimensional study, we count a wall, room or LIO attribution as an answer, only if it has converged and its estimated offset exceeds the BIM-revision threshold. Outputs with no conflict, ambiguous outputs, unavailable positions and failed runs remain in the denominator $N$. The answer rate is $A/N$ and the conditional error rate is $E/A$, where $A$ counts answers and $E$ counts wrong ones~\cite{Chow1970RejectTradeoff, ElYaniv2010Selective}. An answer is counted as correct only when it names the injected fault. The diS-Graphs-style baseline reports walls and rooms but has no LIO attribution. Please refer to \ref{app:lifetest} for the rule governing an additional flag.

On the real sequences, every window is run twice: once unmodified (denoted C0) and once with a single injected fault. A window enters the evaluation only when the unmodified run of our method ends with no conflict. Baselines may still raise structural flags on those windows. Scoring is performed relative to each method's own C0: an output is counted as off-target if it is new with respect to that method's C0 and lies outside the injected fault. For our method, this means off-target attributions; for the baselines, it means off-target structural flags.

Injections cover three types of faults: offsets of single walls, translations of rooms or corridor regions, and translations or rotations of the LIO prior. The magnitudes are the same on all three SLABIM floors. For translational faults they are $\pm0.10$, $\pm0.20$ and $\pm0.30$~m; for LIO prior yaw they are $\pm1^\circ$, $\pm3^\circ$ and $\pm5^\circ$.

Absorption measures how much of an injection is absorbed into the pose. It is the change in pose relative to C0, projected onto the injected direction and divided by the injected magnitude. A value of 1 indicates that the pose fully follows the injection, while 0 indicates no movement. For a wall or region injection, the ratio represents the share of the BIM offset pulled into the pose; for an LIO injection, it represents the share of the prior error that is retained. The trust-BIM comparison is restricted to conditions on which both methods produce a value.

Finally, two coverage quantities are defined. In simulation, the coverage rate is the pointwise rate at which the union interval contains the true pose. On SLABIM, the corresponding rate against the recovered reference is called the reference inclusion rate. In Room 209, the laser tracker provides only an initial pose without timestamps, so neither rate is computed.

\subsection{Statistics}
\label{sec:design-statistics}

The specification of the two-dimensional study was fixed before the formal run. Please refer to \ref{app:lifetest} for details. The main comparison covers strong and weak priors and translations larger than 0.05~m. Rates are macro-averages over cells: each non-empty cell has the same weight, and the conditional error rate is the ratio of the averaged $E/N$ and $A/N$. On the cases labeled detectable but not attributable, our method is compared with the abstaining diS-Graphs-style baseline at the same answer rate.

Confidence intervals are simultaneous 99.58\% intervals over a family of 12 comparisons, obtained from 5,000 paired bootstrap resamples~\cite{Efron1979Bootstrap}. The pre-registered criterion for a clear improvement requires that the conditional error rate on these cases be at least 5 percentage points lower than that of the abstaining baseline, and that the simultaneous interval of the paired risk difference lie entirely below zero. In addition, three guardrails must hold: one on the rate of correct attribution, one on false flags, and one on failed runs. The false-flag guardrail, as fixed in the specification, requires that the Wilson lower bound~\cite{Wilson1927} of the rate of wall or room flags in the absence of any injected fault be at most 1\%.

The real-data results are descriptive rather than confirmatory: each floor contributes only a small number of window blocks, the four windows of Room 209 come from two sequences, and prefixes as well as conditions inside a window are not independent of one another.

\section{Results}
\label{sec:results}

This section reports the experimental results, organized by the five research questions. The two-dimensional study answers RQ1--RQ4, where the ground truth is known. Room 209 and SLABIM then examine whether the same rules appear in real buildings (RQ5), and the SLABIM intervals also speak to RQ4.

\subsection{Two-dimensional study}
\label{sec:results-2d}

\begin{figure}[t]
\centering
\includegraphics[width=\linewidth]{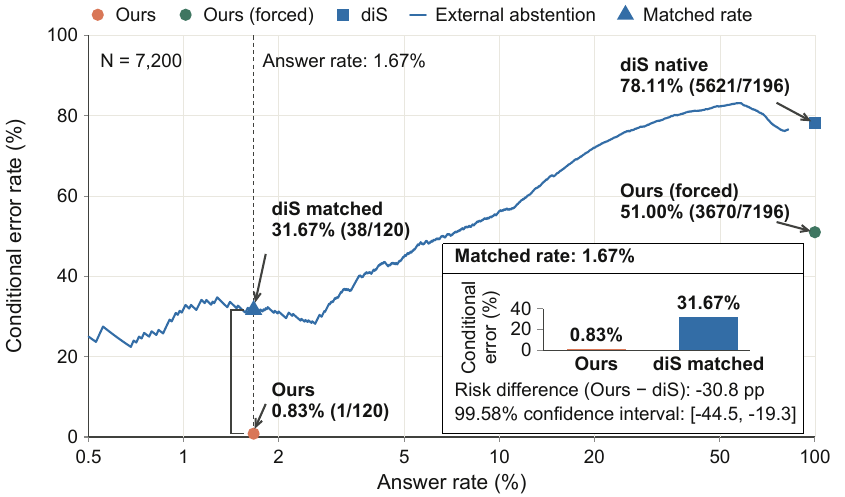}
\caption{Conditional error rate against answer rate on the 7,200 detectable-but-not-attributable cases of the pre-registered two-dimensional study. The line is the diS-Graphs-style baseline with uncertainty-based abstention; markers are native operating points. At the answer rate of Ours (120 of 7,200, 1.67\%), Ours made one wrong attribution (0.83\%) and the abstaining baseline 38 (31.67\%); the paired risk difference is $-30.8$ percentage points (99.58\% CI $-44.5$ to $-19.3$). Restricted to wall faults, the rates are 0.84\% and 31.09\%. Error rates are cell macro-averages.}
\label{fig:risk-coverage}
\end{figure}

Where the cause cannot be told apart, wrong attributions remain rare at the answer rate the method actually uses (RQ1). On the 7,200 cases labeled detectable but not attributable, the conditional error rate of our method at the same answer rate was 0.83\% (1 of 120), against 31.67\% (38 of 120) for the abstaining diS-Graphs-style baseline. The paired risk difference is $-30.8$ percentage points, as shown in Fig.~\ref{fig:risk-coverage}, so the result meets the pre-registered criterion for a clear improvement. Table~\ref{tab:2d-faults} separates wall faults from LIO faults. The baseline has no LIO output, but the gap remains when only walls are counted.

\begin{table*}[t]
\centering
\caption{Pre-registered two-dimensional study by label and fault type (strong and weak priors, translations larger than 0.05~m). $A$: answers; $E/A$: conditional error rate (cell macro-average). Counts are unweighted.}
\label{tab:2d-faults}
\small
\begin{tabular}{@{}>{\raggedright\arraybackslash}p{0.14\textwidth}>{\raggedright\arraybackslash}p{0.13\textwidth}>{\raggedright\arraybackslash}p{0.19\textwidth}>{\raggedright\arraybackslash}p{0.17\textwidth}>{\raggedright\arraybackslash}p{0.25\textwidth}@{}}
\toprule
Label & Faults & Ours & Forced variant & diS-Graphs-style \\
\midrule
Detectable but not attributable & All ($N=7{,}200$) & $A=120$ (1.67\%); $E/A=0.83\%$ & $A=7{,}196$; $E/A=51.00\%$ & Native: $A=7{,}196$, $E/A=78.11\%$; abstaining at the answer rate of Ours: $E/A=31.67\%$ \\
 & Walls ($N=2{,}400$) & $A=119$; $E/A=0.84\%$ & -- & Abstaining at the same answer rate: $E/A=31.09\%$ \\
\midrule
Attributable & All ($N=5{,}600$) & $A=5{,}574$; no wrong attribution & $A=5{,}595$; no wrong attribution & $A=5{,}596$; 2,705 correct (48.30\%) \\
 & Walls ($N=3{,}200$) & 3,200 correct & -- & 2,705 correct (84.53\%) \\
 & LIO ($N=2{,}400$) & 2,374 correct & -- & 0 correct (no LIO output) \\
\midrule
Undetectable & All ($N=16{,}000$) & $A=164$; $E/A=0.61\%$ & $A=1{,}854$; $E/A=44.50\%$ & $A=11{,}443$; $E/A=94.49\%$ \\
\midrule
No fault & $N=3{,}200$ & No wall or room flag & -- & -- \\
\bottomrule
\end{tabular}
\end{table*}

The three guardrails also hold under the pre-registered criterion: the paired difference in the correct-attribution rate on attributable cases lies between 50.3 and 52.1 percentage points, the 3,200 fault-free runs flag no wall or room (Wilson upper bound 0.12\%), and no supported run failed. Where the fault is attributable, our method keeps answering (RQ2). On the 5,600 attributable cases it answered 5,574 times with no wrong attribution, and the forced variant answered 5,595 times, also with no error. Table~\ref{tab:2d-faults} splits wall faults from LIO faults. Most of the baseline's errors name the wrong element, as shown in Table~\ref{tab:dis-errors}. Under an LIO fault it flagged both walls and rooms in 68.1\% of the runs, because it has no explanation for an odometry error and accounts for the localization error as structural movement. Flags of that kind, if written into the BIM, would lead to the movement of walls and rooms that actually stand where the model places them.

\begin{table*}[t]
\centering
\caption{Errors of the diS-Graphs-style baseline in the pre-registered two-dimensional study. Top: outcome of each run by label; strict scoring requires additional flags to meet the 3~cm rule, lenient scoring ignores them. Bottom: flags raised under an LIO fault (14,400 runs) and the magnitude of the flagged offsets (per flagged element).}
\label{tab:dis-errors}
\small
\setlength{\tabcolsep}{3pt}
\begin{tabular}{@{}lrrrrrrr@{}}
\toprule
Label & $N$ & Missed & Wrong primary & Extra only & Correct & Strict & Lenient \\
\midrule
Attributable & 5,600 & 4 & 2,681 & 210 & 2,705 & 48.30\% & 52.05\% \\
Det.\ not attr. & 7,200 & 4 & 5,426 & 195 & 1,575 & 21.88\% & 24.58\% \\
Undetectable & 16,000 & 4,557 & 10,662 & 151 & 630 & 3.94\% & 4.88\% \\
\midrule
\multicolumn{3}{@{}l}{Flags under an LIO fault} & \multicolumn{2}{r}{Runs} & \multicolumn{3}{r}{Share} \\
\multicolumn{3}{@{}l}{Walls and rooms} & \multicolumn{2}{r}{9,807} & \multicolumn{3}{r}{68.10\%} \\
\multicolumn{3}{@{}l}{Rooms only} & \multicolumn{2}{r}{86} & \multicolumn{3}{r}{0.60\%} \\
\multicolumn{3}{@{}l}{Walls only} & \multicolumn{2}{r}{54} & \multicolumn{3}{r}{0.38\%} \\
\multicolumn{3}{@{}l}{None} & \multicolumn{2}{r}{4,453} & \multicolumn{3}{r}{30.92\%} \\
\midrule
\multicolumn{3}{@{}l}{Flagged offset (median; P95)} & \multicolumn{5}{r}{walls 0.16; 0.56~m \quad rooms 0.19; 0.67~m} \\
\bottomrule
\end{tabular}
\end{table*}

The union interval of Section~\ref{sec:method} covers the true pose more often than the baseline's marginal intervals. Pointwise coverage is 94.75--98.78\% across the three labels, against 57.05--84.08\%, as reported in Table~\ref{tab:lifetest-full}. In the demonstration below, between 6\% and 32\% of the trajectories leave the interval at least once, except in the LIO world with a weak prior.

In addition, half the cells of the recovery grid and of the paired-world grid have no later reference, so those cells cannot be resolved, as shown in Table~\ref{tab:recovery}. Our method reaches a correct and stable resolution in 42.22\% of the recovery runs and in 33.55\% of the paired worlds. The forced variant reaches nearly the same share, but its wrong attributions are 27.58\% and 20.86\%, against 0.42\% and 0.19\% for our method. Forcing an answer therefore adds wrong attributions rather than correct ones. The Kaplan--Meier median, measured from first detection and with unresolved runs right-censored, is 8.0~m. Whether a new reference restores the attribution is examined in the demonstration below.

\begin{figure}[t]
\centering
\includegraphics[width=\linewidth]{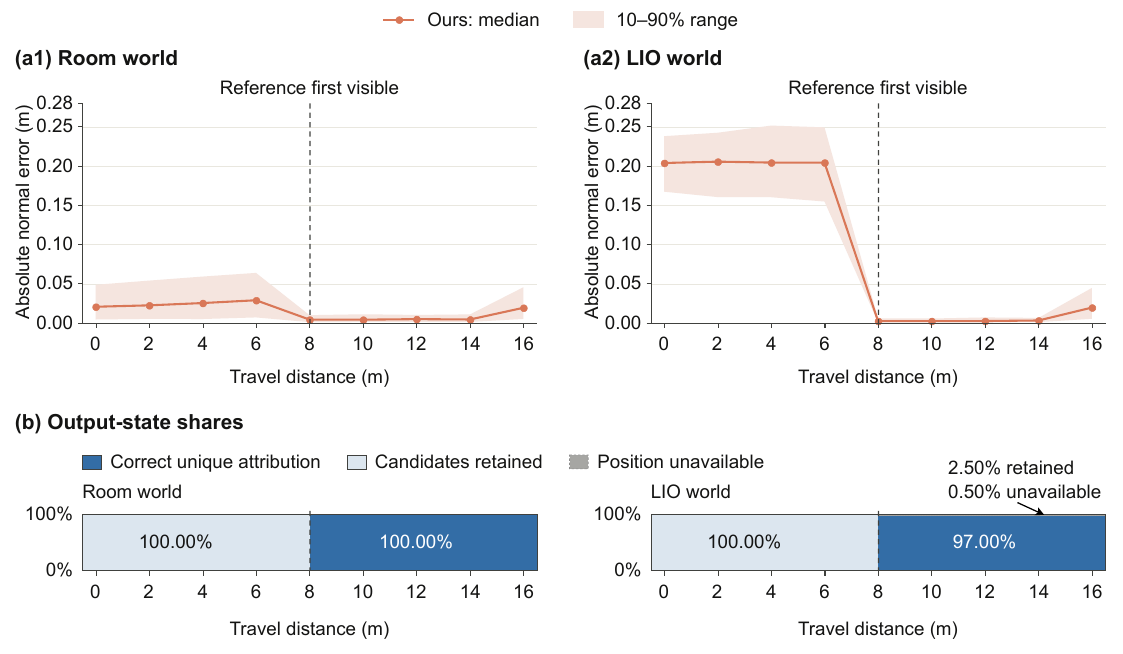}
\caption{Deferred BIM anchoring in the paired worlds with a strong prior ($\sigma=0.03$~m; 200 trials per world). (a) Working-pose error along the ambiguous direction in the room world (a1) and the LIO world (a2): median and 10--90\% range. Before the reference becomes visible, the pose follows odometry in the ambiguous direction, so the LIO world carries the prior offset (RMSE about 0.21~m, against 0.03~m in the room world); once the reference is visible, both errors fall to about zero. (b) Share of trials in each output state. Ambiguity was resolved correctly in 200 of 200 room-world and 194 of 200 LIO-world trials, with no wrong unique attribution.}
\label{fig:deferred}
\end{figure}

\begin{figure}[t]
\centering
\includegraphics[width=\linewidth]{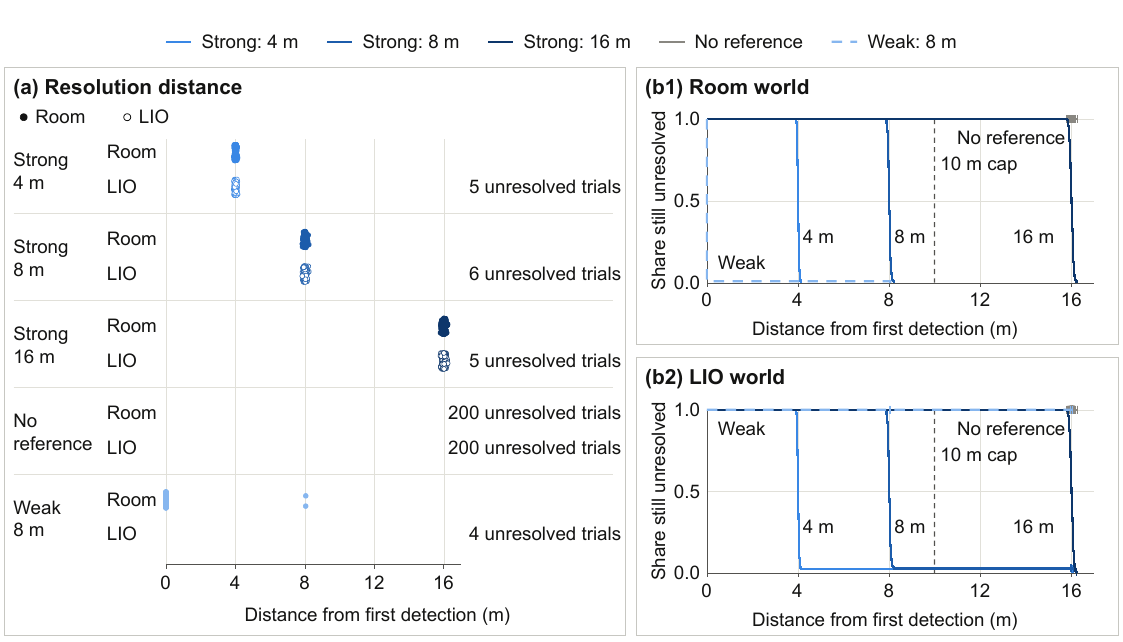}
\caption{Resolution of ambiguity after a new reference appears (200 trials per configuration and world). (a) Resolution distance of each trial, measured from first detection. (b) Kaplan--Meier estimate~\cite{KaplanMeier1958} of the share still unresolved against distance travelled in the room world (b1) and the LIO world (b2); ticks mark censoring and the dashed line marks the 10~m distance cap. With the reference at 16~m, all 200 trials in each world are first withdrawn at the cap; the ambiguity is then resolved in 200 room-world and 195 LIO-world trials. Without a reference, every trial is withdrawn after 10~m. With the weak prior, the LIO offset went undetected in 196 LIO-world trials, where H0 is already correct; these are not in the risk set.}
\label{fig:resolution}
\end{figure}

A separate demonstration follows one pair of worlds with a strong prior, $\sigma=0.03$~m, as illustrated in Fig.~\ref{fig:deferred}. Until the reference wall outside the door is visible, the robot sees one room, both explanations remain, and the ambiguous direction follows odometry. The room world then sits at 0.03~m, while the LIO world carries its prior offset, about 0.21~m. Once the wall is visible, both errors fall to about zero. The ambiguity was resolved in all 200 room-world trials and in 194--195 of 200 LIO-world trials, and no configuration produced a wrong unique attribution. Fig.~\ref{fig:resolution} shows how far the robot travels before the ambiguity ends, including trials in which the reference wall arrives at 16~m or not at all. With the weak prior the room offset went undetected and the position stayed about 0.20~m off until the reference wall appeared; 196 of 200 LIO-world trials raised no conflict, where H0 was already correct.

In a fixed window of 200 trials, every trial was ambiguous at the first keyframe and was resolved inside the window, which corrects the earlier keyframes as well. The error of the first keyframe along the wall normal fell from 29.9~mm to 5.2~mm.

\subsection{Laboratory room 209}
\label{sec:results-209}

\begin{figure*}[t]
\centering
\includegraphics[width=0.8\linewidth]{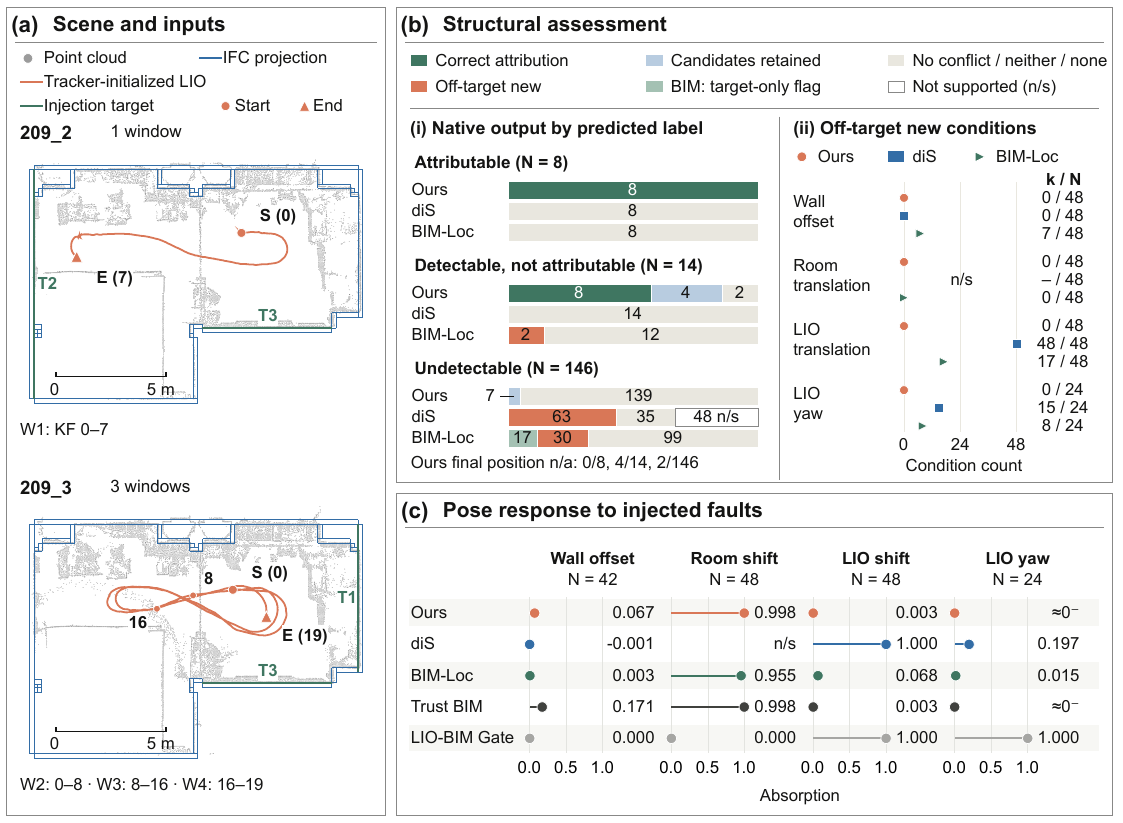}
\caption{Laboratory room 209 with its IFC model; both test sequences start from the laser-tracker initial pose. (a) Point cloud over the IFC walls, with the trajectories, the four windows and the injection targets. (b) Native outputs by predicted label, and off-target attributions or structural flags by fault type, over 168 injected conditions. Ours attributed all 8 predicted-attributable faults correctly and raised no off-target attribution; the diS-Graphs-style and BIM-Loc-style baselines raised off-target structural flags in 63 and 32 conditions. Of the 14 detectable-but-not-attributable conditions, Ours attributed 8 correctly, retained candidates in 4 and found no conflict in 2; four of these six ended with the position unavailable. The diS-Graphs-style baseline flagged 7--14 structures in each of these 14 conditions, none of them a new off-target flag relative to its C0. (c) Median absorption by fault type on the conditions shared with Ours. Trust BIM and the LIO-BIM-style gate report no attributions or structural flags, so they appear only in (c). In a single room, a room translation and an LIO translation form paired worlds, so every method that corrects the pose with the BIM follows the room translation (0.955--0.998). Ours does not misreport it, and flags $x$ and $y$ as unverified in 23 of its 26 control outputs. The four windows come from two sequences, so the counts are descriptive.}
\label{fig:lab209}
\end{figure*}

Room 209 is the paired-world geometry in a real room: one room and a tracker pose at the start (RQ5). Of 168 injected conditions, 8 were predicted attributable. All 8 were attributed correctly, and the 168 conditions produced no off-target attribution, whereas the diS-Graphs-style and BIM-Loc-style baselines raised off-target structural flags in 63 and 32 conditions, as shown in Fig.~\ref{fig:lab209}. Of the 14 conditions predicted detectable but not attributable, our method attributed 8 correctly, kept the candidates in 4 and found no conflict in 2, with no wrong unique attribution, and 4 of those 6 ended with the position unavailable. Across all 168 conditions, the position was unavailable in only 6 cases.

A room translation was absorbed by 99.8\%, and by 95.5\% under the BIM-Loc-style baseline. The pose is pulled because a single room cannot separate a room translation from an LIO translation; under every method that corrects with the BIM the pose follows the room, just as the weak-prior case in Section~\ref{sec:method-anchoring} follows H0. On the undetectable injections there was no off-target attribution, and $x$ and $y$ were marked unverified on 23 of the 26 unmodified outputs. Wall offsets on the shared subset were absorbed by 6.7\%, against 17.1\% for the trust-BIM ablation, and an LIO translation by 0.3\%.

\subsection{SLABIM}
\label{sec:results-slabim}

\begin{figure*}[t]
\centering
\includegraphics[width=0.8\linewidth]{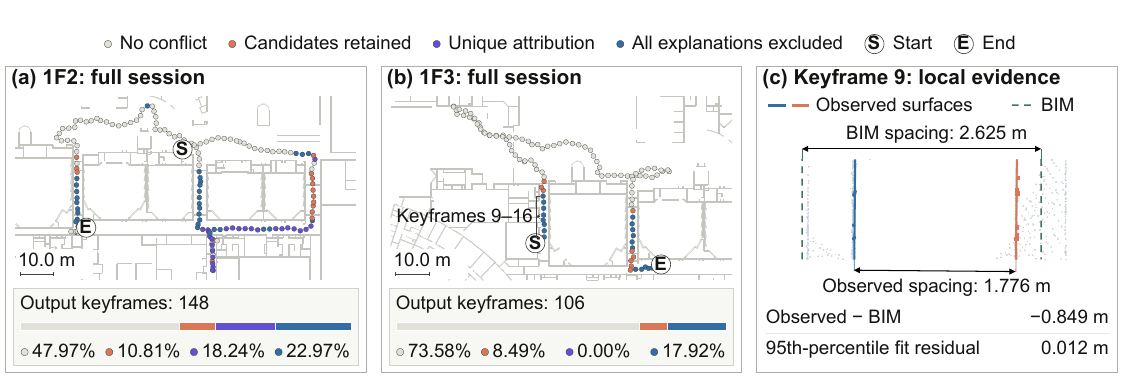}
\caption{Ours on the original BIM over two full sessions of SLABIM floor 1F, the development floor. (a) Session 1F2 (148 outputs) and (b) session 1F3 (106 outputs), with keyframes colored by output state; unique attributions cannot be verified on the original BIM. (c) Keyframe 9 of 1F3: the two observed surfaces associated with walls 619 and 6459 are 1.776~m apart, against 2.625~m in the BIM. Where the scan cannot be reconciled with the BIM, Ours withdraws the position rather than forcing an answer. The recovered references disagree by about 18~cm between the sessions, so the states are descriptive.}
\label{fig:sessions-1f}
\end{figure*}

On the original BIM of floor 1F, the development floor, the position is withdrawn where no single fault fits the scan, as illustrated in Fig.~\ref{fig:sessions-1f}. Our method excluded all explanations in 22.97\% of the outputs of session 1F2 and in 17.92\% of session 1F3. The exclusions cluster in 10 segments, where in several places the opposite surfaces are 0.04--1.1~m closer together than the BIM walls. In five places, the lower part of a surface deviates while the upper part lies near the BIM. 

Moreover, the 27 unique attributions on session 1F2 were left as proposals, and 15 of the wall estimates sit at the 0.35~m bound. The position was available in 111 of 148 outputs on 1F2 and in 87 of 106 on 1F3. Near 55~m of 1F3, the output reports no conflict, but the heading differs from the recovered reference by about 8$^\circ$, against about 1$^\circ$ for FAST-LIO2 alone. In the second half of 1F2, the output lies about 1~m from the recovered reference, against about 4~m for FAST-LIO2 alone.

\begin{figure}[t]
\centering
\includegraphics[width=\linewidth]{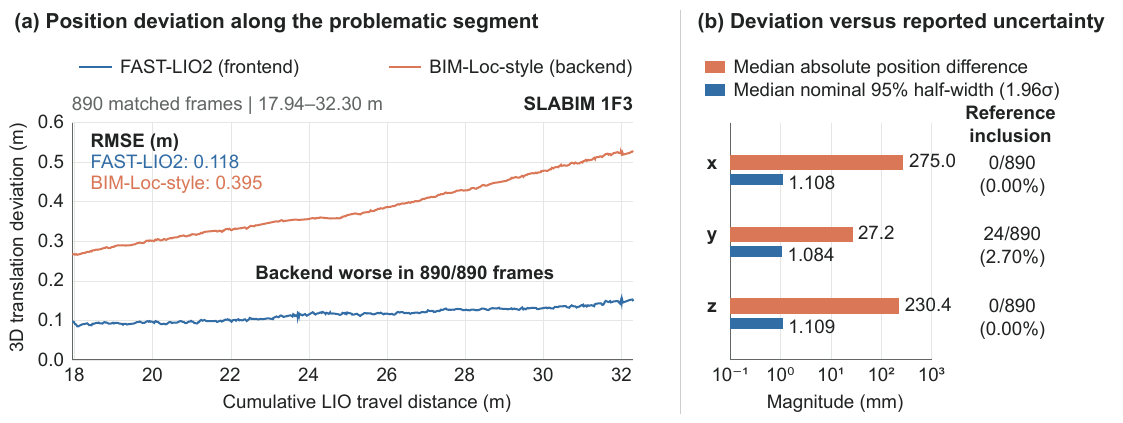}
\caption{The 1F3 segment where Ours withdraws its position (keyframes 9--16, about 18--32~m, 890 frames). (a) Translation error against the recovered reference: the BIM-Loc-style backend (RMSE 0.395~m) is worse than FAST-LIO2 alone (0.118~m) in every frame. (b) Absolute error and self-reported 95\% half-width per axis, log scale. The median per-axis $\sigma$ is about 0.56~mm, and the nominal 95\% intervals include the reference in 0\% ($x$, $z$) and 2.7\% ($y$) of the frames. The reproduction uses an odometry prior 8--10 times tighter than measured. A covariance computed on the assumption that the map is correct cannot reflect errors in the map.}
\label{fig:bimloc-segment}
\end{figure}

On the segment where the position is withdrawn, the BIM-Loc-style baseline keeps localizing against the BIM, as shown in Fig.~\ref{fig:bimloc-segment}. Over keyframes 9--16 of session 1F3, about 18--32~m, its translation RMSE is 0.395~m, against 0.118~m for FAST-LIO2 alone, and its error is larger on every one of the 890 frames. The median of its self-reported per-axis standard deviation on that segment is about 0.56~mm, which illustrates the same gap noted in Section~\ref{sec:intro}: a covariance computed on the assumption that the map is correct cannot reflect an error in the map.

On the keyframes where the position was withdrawn, the BIM-Loc-style three-dimensional RMSE against the recovered reference is 0.625~m on 1F2 (37 keyframes) and 1.252~m on 1F3 (19 keyframes), against 0.153~m and 0.191~m for FAST-LIO2. These are three-dimensional differences from the recovered reference, not errors of our planar working pose. Over the full sessions, the backend lies 0.694~m and 1.247~m from the recovered reference, against 0.157~m and 0.257~m for FAST-LIO2. 

\begin{figure*}[t]
\centering
\includegraphics[width=0.8\linewidth]{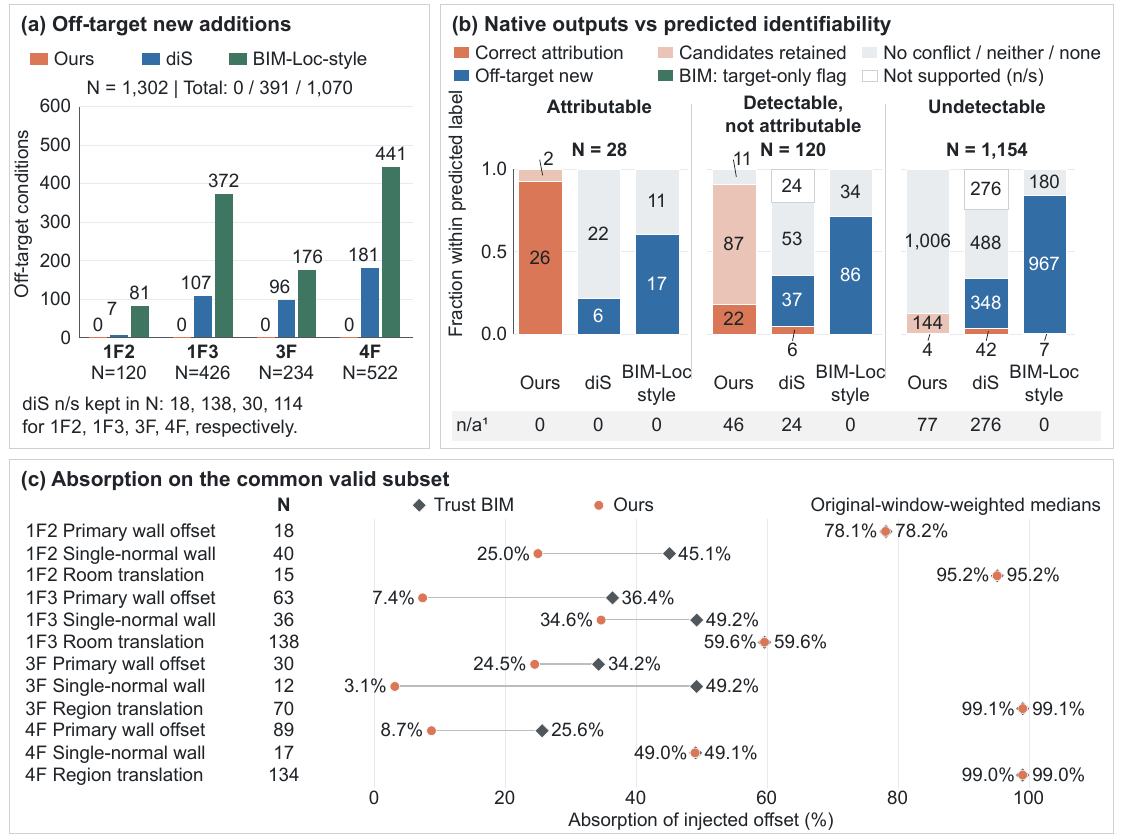}
\caption{Paired fault injections on SLABIM floors 1F, 3F and 4F (1F is the development floor). (a) Conditions with an off-target attribution (Ours) or off-target structural flag (baselines), counted as new relative to each method's own C0, by floor. Ours raised none in 1,302 conditions; the two baselines raised 391 and 1,070. (b) Native outputs by predicted label; BIM-Loc-style flags are not causal attributions. (c) Median absorption of the injected offset into the pose for the trust-BIM ablation and the working pose of Ours, on the conditions where both are available. Region injections are geometrically degenerate and form paired worlds, so both absorb them almost fully. Window blocks are few, so the results are descriptive.}
\label{fig:slabim-injections}
\end{figure*}

Across the 1,302 injected conditions, our method proposed no BIM change outside the injected fault: it raised no off-target attribution, while the diS-Graphs-style and BIM-Loc-style baselines raised off-target structural flags on 391 and 1,070, as shown in Fig.~\ref{fig:slabim-injections} and Table~\ref{tab:injections}.

\begin{table*}[t]
\centering
\caption{Paired injections on the real data, by floor and fault type ($N$ conditions). Off-target: conditions with an off-target attribution (Ours) or off-target structural flag (baselines), new relative to each method's own C0; a condition that a baseline does not support counts as having no flag, and the diS-Graphs-style baseline supports no room translation in room 209. Absorption: median share, in \%, of the injected offset that the pose follows; the trust-BIM ablation is evaluated on the conditions where Ours has a value, the other methods on their own valid conditions, with the number of conditions in parentheses when it is smaller than $N$. Reference inclusion: conditions whose interval includes the recovered reference; not available in room 209. Room 209 pools its two test sequences, 209\_2 (one window) and 209\_3 (three windows). Region translations on 4F move a corridor region. 1F is the development floor.}
\label{tab:injections}
\scriptsize
\setlength{\tabcolsep}{2.5pt}
\begin{tabular}{@{}ll r rrr rrrrr rrr@{}}
\toprule
 & & & \multicolumn{3}{c}{Off-target} & \multicolumn{5}{c}{Absorption (\%)} & \multicolumn{3}{c}{Reference inclusion} \\
\cmidrule(lr){4-6}\cmidrule(lr){7-11}\cmidrule(l){12-14}
Floor & Fault & $N$ & Ours & diS & BIM-Loc & Ours & Trust BIM & diS & BIM-Loc & Gate & Ours & Trust BIM & BIM-Loc \\
\midrule
Room 209 & Wall offset & 48 & 0 & 0 & 7 & 6.7 (42) & 17.1 (42) & -0.1 & 0.3 & 0.0 & -- & -- & -- \\
 & Room or region translation & 48 & 0 & 0 & 0 & 99.8 & 99.8 & -- & 95.5 & 0.0 & -- & -- & -- \\
 & LIO prior translation & 48 & 0 & 48 & 17 & 0.3 & 0.3 & 100.0 & 6.8 & 100.0 & -- & -- & -- \\
 & LIO prior yaw & 24 & 0 & 15 & 8 & 0.0 & 0.0 & 19.7 & 1.5 & 100.0 & -- & -- & -- \\
\midrule
1F2 (dev.) & Wall offset & 18 & 0 & 0 & 6 & 78.1 & 78.2 & 1.5 & 0.0 & 0.0 & 18/18 & 8/18 & 0/18 \\
 & Wall offset, single normal & 42 & 0 & 7 & 16 & 25.0 (40) & 45.1 (40) & 0.0 & 0.0 & 0.0 & 36/42 & 21/42 & 0/42 \\
 & Room or region translation & 24 & 0 & 0 & 23 & 95.2 (15) & 95.2 (15) & 0.0 (6) & 0.2 & 0.0 & 24/24 & 9/24 & 0/24 \\
 & LIO prior translation & 24 & 0 & 0 & 24 & 3.0 (16) & 3.0 (16) & 100.0 & 35.2 & 100.0 & 23/24 & 18/24 & 0/24 \\
 & LIO prior yaw & 12 & 0 & 0 & 12 & 0.4 & 0.4 & 6.7 & 40.1 & 100.0 & 12/12 & 6/12 & 0/12 \\
\midrule
1F3 (dev.) & Wall offset & 72 & 0 & 14 & 56 & 7.4 (63) & 36.4 (63) & 0.0 & 0.0 & 0.0 & 70/72 & 41/72 & 4/72 \\
 & Wall offset, single normal & 36 & 0 & 5 & 23 & 34.6 & 49.2 & 0.0 & 0.0 & 0.0 & 33/36 & 17/36 & 0/36 \\
 & Room or region translation & 156 & 0 & 1 & 134 & 59.6 (138) & 59.6 (138) & 0.0 (18) & 1.8 & 0.0 & 149/156 & 56/156 & 2/156 \\
 & LIO prior translation & 108 & 0 & 78 & 107 & 0.4 & 0.4 & 100.0 & 11.8 & 100.0 & 108/108 & 60/108 & 2/108 \\
 & LIO prior yaw & 54 & 0 & 9 & 52 & 0.1 & 0.1 & 0.4 & 44.6 & 100.0 & 42/54 & 35/54 & 0/54 \\
\midrule
3F & Wall offset & 42 & 0 & 20 & 25 & 24.5 (30) & 34.2 (30) & 0.0 & 0.1 & 47.9 & 39/42 & 19/42 & 4/42 \\
 & Wall offset, single normal & 12 & 0 & 10 & 5 & 3.1 & 49.2 & 0.0 & 1.2 & 49.2 & 12/12 & 2/12 & 0/12 \\
 & Room or region translation & 72 & 0 & 0 & 59 & 99.1 (70) & 99.1 (70) & 0.0 (42) & 1.0 & 99.1 & 72/72 & 12/72 & 3/72 \\
 & LIO prior translation & 72 & 0 & 66 & 60 & 0.6 & 0.6 & 100.0 & 0.4 & 0.9 & 72/72 & 66/72 & 3/72 \\
 & LIO prior yaw & 36 & 0 & 0 & 27 & 0.2 & 0.2 & 0.1 & 1.1 & 0.0 & 36/36 & 36/36 & 4/36 \\
\midrule
4F & Wall offset & 108 & 0 & 33 & 69 & 8.7 (89) & 25.6 (89) & 0.0 & 0.0 & 9.0 & 91/108 & 55/108 & 0/108 \\
 & Wall offset, single normal & 24 & 0 & 15 & 13 & 49.0 (17) & 49.1 (17) & 0.0 & 0.0 & 33.6 & 23/24 & 5/24 & 0/24 \\
 & Room or region translation & 156 & 0 & 3 & 145 & 99.0 (134) & 99.0 (134) & 0.0 (42) & 1.6 & 50.6 & 152/156 & 36/156 & 0/156 \\
 & LIO prior translation & 156 & 0 & 126 & 143 & 0.4 & 0.4 & 100.0 & 0.7 & 100.0 & 156/156 & 103/156 & 5/156 \\
 & LIO prior yaw & 78 & 0 & 4 & 71 & 0.0 & 0.0 & 0.3 & 0.3 & 100.0 & 59/78 & 56/78 & 2/78 \\
\bottomrule
\end{tabular}

\smallskip
\footnotesize diS: diS-Graphs-style; BIM-Loc: BIM-Loc-style; Gate: LIO-BIM-style gate.
\end{table*}

On floor 4F, where Section~\ref{sec:identifiability} predicts that a well-witnessed wall can be attributed, the true explanation remained among the candidates in all 522 conditions. Seventeen of the 18 attributable wall offsets were attributed, and the other stayed ambiguous. Wall offsets were absorbed by a median of 8.7\% on the conditions shared with the trust-BIM ablation, against 25.6\% for that ablation, and LIO prior translations by 0.4\%, against 100\% for the diS-Graphs-style baseline. On 4F, the single-normal walls, those with no third wall along the offset direction, are absorbed by 49.0\% into the working pose and by 49.1\% under the trust-BIM ablation. The union interval includes the recovered reference in 23 of the 24 conditions, against only 5 of 24.

A translation along a 4F corridor is a paired world in the real building. A corridor is not a stand-in for an ordinary room, but both the working pose and the trust-BIM ablation absorb the translation by 99.0\%. Among the no-conflict outputs that absorbed most of it, 84 of 97 mark the corridor direction unverified. In one condition, the region moved 0.10~m along $-y$ and the final pose moved $-0.0998$~m, with $y$ marked unverified. In one 3F condition, a single-normal wall was moved by $-0.20$~m; the working pose absorbed 1.0\% of the offset and the trust-BIM ablation 49.2\%, so the overlay showed about 0.198~m against about 0.101~m.

On floor 3F, the 234 conditions produced no off-target attribution, and the faults predicted attributable were attributed. Single-normal walls were detected and the candidates were kept, which brings absorption down to 3.1\%, against 49.2\% for the trust-BIM ablation. With only one ordinary room, and with the reference wall outside it appearing on the last keyframe of the window, that room was not resolved. The two resolved cases among the 72 room and region conditions are corridor regions.

On floor 1F, 120 conditions of session 1F2 and 426 of session 1F3 produced no off-target attribution (Table~\ref{tab:injections}). Please refer to \ref{app:frontend} for the outcomes by predicted label. The main walls of 1F2 cannot be told apart, which leaves the offset absorbed by 78.1\% and by 78.2\% under the trust-BIM ablation. Even so, the union interval includes the recovered reference in all 18 of those conditions, against 8 of 18 for the ablation. Unverified directions are a routine output on these floors. With no conflict, $x$ is unverified in 81 of 119 outputs on 4F and $y$ in 75 of 119, and on 3F $y$ is unverified in 54 of 55. Yaw, by contrast, has support from the BIM on 4F, on 3F and in room 209.

The LIO-BIM-style gate decides whether the BIM is used, not which side of the conflict is wrong. At the thresholds of the official source code, it accepts about 74--79\% of the frames on 3F and 3--8\% on 1F. On 3F a wall offset is then absorbed by 47.9\%; on 1F, the absorption is zero, and the LIO error is kept in full. Stricter thresholds move every floor toward odometry alone, and looser thresholds move the well-matched floors toward the trust-BIM ablation. Please refer to \ref{app:baselines} for details.

\subsection{Runtime}
\label{sec:results-runtime}

All timing measurements were performed on an i9-13980HX platform using a single thread, without a GPU. The replay of the fixed inputs takes a median of 35--45~ms per keyframe for the front end and the back end together on SLABIM, and 88~ms in room 209. The slowest keyframe, 214~ms in room 209, used about one sixth of the 1.33~s available at 1.5~m/s. The replay feeds the frozen front end with the archived poses and omits FAST-LIO2, the construction of the geometry and the initialization on the first nine keyframes.

\section{Discussion}
\label{sec:discussion}

\subsection{Engineering implications}
\label{sec:discussion-engineering}

The outcomes of the tests translate into different actions on site. Each row of Table~\ref{tab:update} corresponds to one such action. A conflict and its candidates are monitoring data, and a revision enters the model only after a person has checked it; this is the separation between monitoring data and model knowledge discussed in Section~\ref{sec:related-scanvsbim}.

On the robot, an LIO bias is handled by correcting the working pose. A wall-surface offset, by contrast, needs a person: the robot stops anchoring on that wall and proposes a revision of the form ``the surface observed at wall $w$ deviates from the BIM by about $\delta$,'' and the inspector either revises the as-built model or records an object on site. In session 1F2, 15 of the 20 wall attributions reach the 0.35~m bound, and some of them sit where the session also shows a surface departing at the bottom and meeting the BIM higher up.

A room translation is also a proposal, and it is relative to the datum. Before the room is revised, the start region that carries the datum has to be checked against the building. While the cause is ambiguous the robot keeps moving: directions on which the candidates agree stay anchored to the BIM, and ambiguous directions follow odometry until a new witness resolves them or the distance cap withdraws the coordinate. At that moment, the operator is left with the candidates and the union interval.

When all explanations are excluded, the measurements fit no single fault. Withdrawing the position and proposing nothing leaves the segment as a place to inspect. On floor 1F, those segments are the ones reported in Section~\ref{sec:results-slabim}. With no conflict, the working pose uses the BIM, and a direction marked unverified has not been checked. An overlay of the map on the BIM hides a deviation along that direction, because the pose absorbs it. The corridor condition of Section~\ref{sec:results-slabim}, a 0.10~m translation along $-y$ absorbed into the pose and flagged, is the example. Progress and quality checks that compare a scan with the BIM have to treat a flagged direction as unchecked.

\subsection{Localization based on the BIM and fusion with it}
\label{sec:discussion-stance}

BIM-based localization measures the robot against the model, and gating decides when, or which part of, that reference may be used~\cite{Stuehrenberg2025LIOBIM, Zhang2026BIMLoc}. From floor to floor, the LIO-BIM-style gate in Section~\ref{sec:results-slabim} follows the quality of the match. The fusion here treats the BIM as one more source of evidence, tested at every conflict alongside the odometry. The cost of an error changes with that choice: where no single fault explains the disagreement, the position is withdrawn, while a method that takes the map as correct keeps a confident estimate on the same segment.

This comes at the cost of availability. On the original BIM of floor 1F the position was available in 111 of 148 outputs and in 87 of 106, and on the keyframes where it was withdrawn the BIM-Loc-style backend was 0.625~m and 1.252~m from the recovered reference.

\subsection{Limitations}
\label{sec:discussion-limitations}

\emph{Model scope.} The model allows one fault at a time, so measurements that require two faults at once end in a withdrawn position and call for inspection. On the original BIM of floor 1F the exclusions are those of Section~\ref{sec:results-slabim}. Wall-surface offsets larger than 0.35~m lie outside the model. Because the walls are projected to the plane and taken to be vertical, floors, ceilings and inclined surfaces are not used, and an attribution is a horizontal offset. Telling an as-built deviation of a wall from an object in front of it requires a witness that the wall plane does not provide (Section~\ref{sec:discussion-engineering}). The threats to LiDAR positioning in the field are wider than these three classes~\cite{Rife2025LidarIntegrity}.

\emph{Datum and initialization.} On SLABIM the first nine keyframes align the trajectory and are not tested, so a wall-surface offset in that segment is not reported, and an offset of the whole start region becomes part of the datum. The datum is propagated through the window priors and becomes less certain with distance; a room translation is therefore reported relative to this datum.

\emph{Tests and intervals.} The $\chi^2$ distributions used in the tests are derived under idealized assumptions. On real sequences, nonlinearity, bounds and model mismatch make them approximations. The pose intervals are weighted least-squares intervals conditioned on the selected explanations, pointwise in their nominal level. Associations are frozen before the tests, so the explanations stay the three fault classes of Section~\ref{sec:problem}~\cite{Joerger2025InnovationProjections}. An offset below the detection threshold can pull the pose while the output still reports no conflict, as in the 8$^\circ$ heading on session 1F3. Thirteen corridor outputs on 4F had a weak constraint along the corridor and were left unmarked; their intervals, 0.32--0.65~m wide, include the reference.

\emph{Deferred anchoring and recovery.} If the LIO is wrong, then until the cause is attributed the error in an ambiguous direction is about the difference between the candidates, together with the random noise. The 10~m cap was chosen for the simulation noise model, and it is not a guarantee of safety. Choosing the lowest-cost explanation along the directions where the candidates agree was not given its own experiment. Recovery is completed inside the window that sees the new reference, and a cause is not handed to the next window. Recovery after passing a door has so far been shown only in simulation. On SLABIM, the reference wall outside the one ordinary room appears only on the last keyframe, and that room was not resolved (Section~\ref{sec:results-slabim}).

\emph{Data and evaluation.} The real sequences come from three floors of one university building and from one laboratory room. Both external baselines are reproductions, and their results only represent the class of method they implement. Please refer to \ref{app:baselines} for details.

\subsection{Future work}
\label{sec:discussion-future}

Several extensions follow from the limitations above. A witness other than the wall plane could separate the two sources of a wall-surface offset. Layering by height is one candidate: on floor 1F the lower part of a surface sometimes deviates while the upper part agrees with the BIM. Semantic cues are another, and a model that admits several faults at once could turn some of the withdrawn segments into attributions. A floor on which a reference outside a room becomes visible in time would carry the recovery result of the simulation onto real data.

\section{Conclusion}
\label{sec:conclusion}

When to trust the BIM depends on whether the data can say what is wrong. A scan that disagrees with the model may come from the odometry pose, from an offset wall surface, or from a shifted room. Which of these can be told apart is set in principle by the geometry of the witnesses, and in a given window by the size of the offset relative to the noise. The BIM can be trusted where the geometry makes the fault attributable and the offset is large relative to the noise. There the method corrects the pose and proposes any change of a wall or a room for a person to check.

Where the explanations cannot be told apart, the BIM is left unchanged, the candidates are kept, and BIM anchoring is deferred in the ambiguous directions until a new witness appears. Where no single fault explains the data, the BIM is left unused and the position is withdrawn. With no conflict, the BIM is confirmed only along the directions that independent witnesses constrain; the other directions are reported as unverified, and a use that compares the scan with the BIM has to treat them as unchecked.

In the pre-registered two-dimensional study, the conditional error rate on cases that are detectable but not attributable was 0.83\%, against 31.67\% for a diS-Graphs-style baseline abstaining by its own uncertainty, at the same answer rate. The method answered 5,574 of the 5,600 attributable cases with no wrong attribution. In paired injections on a laboratory room with its IFC model and on three floors of SLABIM, it raised no off-target attribution in 168 and 1,302 conditions, respectively, while both reproduced baselines raised off-target structural flags in every setting.

\appendix

\section{Proof of Proposition~\ref{prop:nesting}}
\label{app:proof}

\begin{proof}
For a fixed bias $b$, minimizing $\lVert\bar r+A\Delta X+C_hb\rVert^2$ over $\Delta X$ leaves $\lVert P^\perp(\bar r+C_hb)\rVert^2$.
Because $\mathcal{B}_h$ is symmetric about zero, minimizing over $b\in\mathcal{B}_h$ gives $J_h=\min_{e\in E_h}\lVert P^\perp\bar r-e\rVert^2=\operatorname{dist}(P^\perp\bar r,E_h)^2$, which is Eq.~\eqref{eq:cost-distance}.
(a) If $E_h\subseteq E_j$, the distance of any point from $E_j$ is at most its distance from $E_h$, so $J_j\le J_h$; an observation with $J_h=0$ also has $J_j=0$.
(b) $E_h$ and $E_j$ are closed, as a subspace or a segment of a line, so in the noise-free case an explanation explains the observation exactly when $P^\perp\bar r$ lies in its set. If $P^\perp\bar r\in E_j\setminus E_h$, then $J_j=0$ and $J_h>0$.
\end{proof}

\section{Baseline reproduction}
\label{app:baselines}

\emph{diS-Graphs-style baseline.} The diS-Graphs-style baseline is a reproduction of diS-Graphs~\cite{Shaheer2025DiSGraphs}. It evaluates the wall deviation in a local frame at the midpoint of the finite wall segment, rebuilds rooms relative to the wall center lines, and takes the direction at zero from the normal. These three steps correct coordinate errors of our first reconstruction, which appears only in Tables~\ref{tab:lifetest-full} and~\ref{tab:recovery}; the errors were in our reconstruction, not in the authors' implementation. It keeps the model nodes, the soft weights and all numerical parameters of the original. Under four coordinate shifts its results agree to within $1.3\times10^{-7}$~m, and its release tests with wall offsets of 0.10, 0.20 and 0.30~m succeeded in 25 of 25 runs each. Uncertainty-based abstention scores each output by the largest standard deviation among its flagged walls and the square root of the largest eigenvalue of the covariance of each flagged room; an output without a finite, positive semidefinite marginal covariance is withheld at every finite threshold. The comparison at the answer rate of our method mixes the two neighboring thresholds at random. The mixture is part of the evaluation and is not offered as a separate method.

\emph{BIM-Loc-style baseline.} The reproduction of BIM-Loc~\cite{Zhang2026BIMLoc} is evaluated on sessions that were not used for development, with no parameter tuning. On the held-out session 3F-Region1, its translation ATE is 0.157~m (published: 0.147~m), its rotation error 0.617$^\circ$ (0.748$^\circ$) and its scan-to-BIM RMSE 0.066~m (0.052~m), within the preset tolerances; FAST-LIO2 alone reaches 0.141~m on the same session. On 4F-Region1, its translation ATE is 0.280~m (published: 0.098~m; preset tolerance: 0.147~m), against 0.512~m for FAST-LIO2 alone. Possible reasons for the difference include the different front end (DLO~\cite{Chen2022DLO} in the original experiments, FAST-LIO2 here), unpublished parameters, and geometric approximations. In the parameter combination used here, the hit threshold and the ray extension are both 0.20~m, the inconsistency of a patch is the mean of its anchor Beta means, and the threshold is 0.85. A BIM patch that lies more than 0.20~m farther from the sensor than the actual wall then receives no evidence and stays undecided; a patch offset toward the sensor receives many misses, but mixed with hits the evidence may not reach the threshold. In a capability test on 4F, offsets of 0.5 and 1.0~m of well-observed walls were not flagged. On the real data, the BIM-Loc-style baseline starts from the recovered reference with an initial standard deviation of about 2~mm. The odometry prior of the reproduction is 8--10 times tighter than measured.

\emph{LIO-BIM-style gate.} The official LIO-BIM implementation requires orientation from the IMU and a BIM in IFC format, whereas SLABIM provides a six-axis IMU without orientation and a mesh BIM, and the IMU of the laboratory platform is six-axis as well. The gate reproduces the decision of LIO-BIM~\cite{Stuehrenberg2025LIOBIM} to use the BIM only after a good match. It fits all walls, computes the match metrics per frame at that fit, keeps the frames that pass, and refits with the walls of these frames only. The main thresholds follow the official source code: a mean squared nearest-neighbor distance of at most 0.1~m$^2$, which the code calls RMSE, a fitness of at least 0.65 with all eligible points as the denominator, and correspondences within 0.6~m$^2$. The paper defines the RMSE threshold as 0.1~m, that is, 0.01~m$^2$ in squared distance; this setting is one of the 15 threshold settings of the sensitivity grid. The gate differs from the original in six ways:
\begin{enumerate}
\item it has no edge and planar features;
\item its BIM comes from PLY and DXF files rather than from features generated from IFC;
\item the official environment could not be built;
\item the IMU orientation input is incompatible;
\item convergence is replaced by the convergence flag of the frozen fit;
\item the gate decides in two passes per window instead of online per frame.
\end{enumerate}

Under the strictest settings of the sensitivity grid, squared-distance thresholds of 0.01 and 0.025~m$^2$, the gate rejected every injected frame and never used the BIM; with 0.05~m$^2$, it passed only 0.94\% of the frames on 3F. Only the two loosest thresholds, 0.10 and 0.20~m$^2$, let it pass clearly more frames on 3F than on 1F. The low pass rates on 1F and in room 209 may partly stem from counting all eligible points, clutter included, in the fitness of the gate.

\section{Pre-registered two-dimensional study: specification and additional tables}
\label{app:lifetest}
\setcounter{table}{0}
\setcounter{figure}{0}

This appendix records the sealed specification of the two-dimensional study and the additional tables that support the main results. The specification was sealed before the formal run. The study was run before the working-pose rule for ambiguous directions (Section~\ref{sec:method-anchoring}) was revised. The revision changes neither the attributions nor the candidates nor the intervals, and replaying the main grid with the revised method gives the same flags in every run, so the results apply to the method of Section~\ref{sec:method}. The study asks whether keeping the candidates lowers the number of wrong attributions at the same answer rate when a wall offset, a room translation and an LIO error are hard to tell apart, and whether correct attributions and recovery remain when they can be told apart. It is a two-dimensional, single-fault study of the decision mechanism, with wall identities given; complete three-dimensional systems are outside its comparison.

\emph{Scenes and sensors.} The four layouts are a corridor of width 2.4~m, two perpendicular walls meeting at a corner, a 6~$\times$~4~m room with a door in its east wall, and the same room with a square pillar of 0.4~m. Each layout has references that come into view later: a pillar in the corridor and the corner layouts, and two fixed walls outside the door of the room. Scans have 720 beams over 360$^\circ$ with a range of 12~m and occlusion; the range noise is 0.015~m, and each physical surface has a normal tolerance drawn from $\mathcal{N}(0,0.005^2)$~m$^2$ and shared over the sequence. The start pose of each run is jittered uniformly by up to 0.12~m per axis and 0.08~rad in yaw. Every window has nine keyframes at nominal distances of 0, 2, \dots, 16~m; each 2~m odometry edge consists of two 1~m increments with independent noise. The prior has a translation standard deviation of 0.03 or 0.30~m and a yaw standard deviation of 1$^\circ$, and its mean is the true first pose plus the declared fault and independent noise; runs without a prior inject no prior fault.

\begin{figure}[t]
\centering
\includegraphics[width=\linewidth]{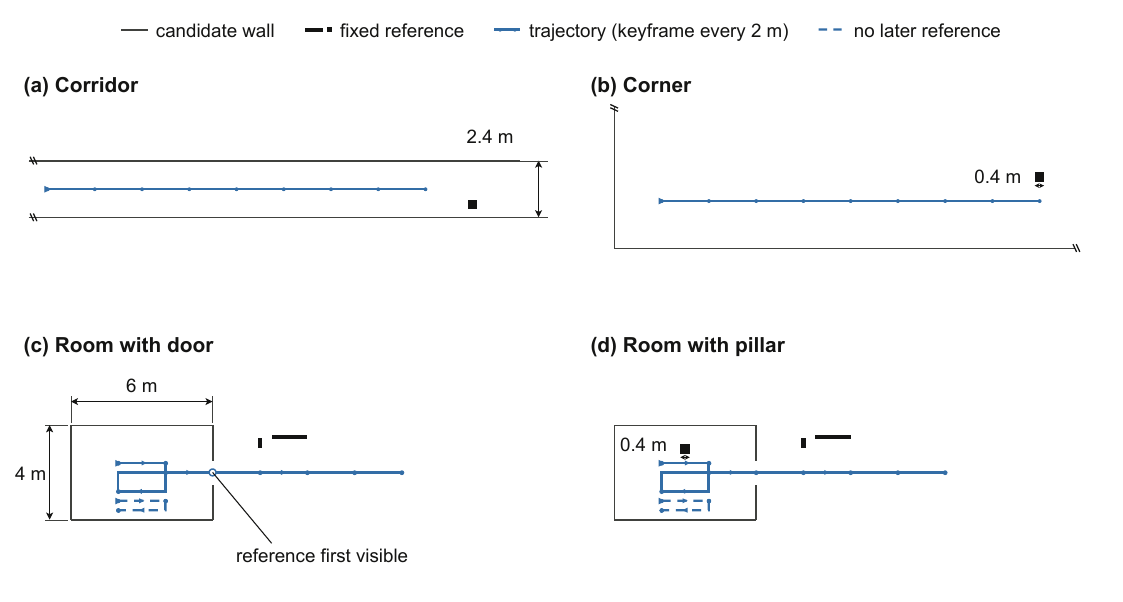}
\caption{Four layouts of the two-dimensional simulation study. (a) Corridor (width 2.4~m). (b) Corner. (c) Room with a door (6~$\times$~4~m). (d) Room with a pillar. Black squares and thick black segments are fixed references; the blue line is the trajectory (keyframe every 2~m); the dashed segment indicates no later reference.}
\label{fig:sim-layouts}
\end{figure}

\emph{Test grid.} Every cell has 400 repetitions, enough to bound a zero error rate below about 1\% by the Wilson upper limit, and the sign of the fault alternates between repetitions. The main grid has 123 cells (49,200 runs): fault-free runs for four layouts and three prior conditions; single-wall offsets of 0.05, 0.10, 0.20 and 0.30~m in three layouts; room translations of 0.10, 0.20 and 0.30~m in three directions; and LIO translations of the same sizes and LIO yaw errors of 1$^\circ$, 3$^\circ$ and 5$^\circ$ in the two room layouts. Wall offsets of 0.05~m lie at the threshold for proposing a revision; they are reported but do not enter the main comparison. The recovery grid has 36 cells (14,400 runs) with and without a reference that comes into view, a diagnostic grid adds a nearly uninformative prior to 25 cells without a prior (10,000 runs), and the paired-world grid has 24 cells of 400 pairs (19,200 worlds). In each pair, the two worlds have identical BIM, scans, associations, priors and odometry until the reference outside the door comes into view, and their true positions differ by the declared translation; this identity is checked for every pair. In total, the study has 92,800 worlds in 73,200 independent groups of random draws.

\emph{Decision rules.} The labels, the scoring and the criterion are those of Section~\ref{sec:design}. An additional flag beside a correct answer is accepted only if the fault affects that element and its estimate lies within 0.03~m of the truth, and an LIO fault requires an LIO attribution. Confidence intervals come from 5,000 paired bootstrap resamples, stratified by cell, that keep each complete repetition and each pair of worlds together. The comparisons of our method with the native and the abstaining baseline on three labels and two measures, risk and correct-attribution rate, form a family of 12; each interval has the level $1-0.05/12$ (99.58\%), so that the family holds at 95\%. The three guardrails of the criterion require that the lower bound of the paired difference in correct-attribution rate on attributable cases, against the strongest native baseline, is at least $-5$ percentage points; that the Wilson lower bound~\cite{Wilson1927} of the rate of wall or room flags without an injected fault is at most 1\%; and that at most 1\% of the runs on supported inputs fail or do not converge. A comparison at the same answer rate requires at least 1,000 independent cases labeled detectable but not attributable, at least 100 answers of our method on them, and an answer rate that the abstaining baseline can reach; otherwise the evidence is recorded as insufficient. The study is behind if the risk at the same answer rate is at least 5 percentage points higher with an interval lower bound above zero, if the upper bound of the difference in correct-attribution rate is below $-5$ percentage points, or if a guardrail fails; otherwise it is a tie. For the intervals, a missing interval counts as not covered, and the share of trajectories that miss the truth at least once is reported separately from the pointwise coverage.

Table~\ref{tab:lifetest-full} gives the main table with both the micro- and the macro-averaged $E/N$. Table~\ref{tab:ablation} gives the native operating points of the suspect-set ablation over the whole main grid; the ablation removes one class of explanation at a time, the LIO or the room. The answer rates differ, and each row stays at its own answer rate. Without the LIO explanation, yaw faults produced detections and no answer (53 detections and no answer in 4,800 runs). Table~\ref{tab:recovery} gives the outcomes on the recovery and paired-world grids. The recovery grid crosses single-wall offsets, room translations, LIO translations and LIO yaw errors (0.20~m and 3$^\circ$) with the strong and weak priors and with and without a later reference; the paired-world grid crosses room and LIO translations of 0.10, 0.20 and 0.30~m along $x$ and $y$ with the same priors and reference conditions. Each run ends in one of four outcomes, which add up to 100\%. A run counts as a wrong attribution if a wrong unique attribution occurred at any time, even if it was later corrected; a run is correct and stable if, after the first correct attribution, every later keyframe stays correct. Table~\ref{tab:params-sim} lists the parameters that the simulated and real configurations share and those of the simulated configuration.

\begin{table*}[t]
\centering
\caption{Main table of the pre-registered two-dimensional study (strong and weak priors, translations larger than 0.05~m). $N$, $A$ and $E$ are unweighted counts; macro-averages weight the cells equally. The rows of the first reconstruction are descriptive and not part of the pre-registered comparison.}
\label{tab:lifetest-full}
\small
\begin{tabular}{@{}l>{\raggedright\arraybackslash}p{0.16\textwidth}rrrrrrr@{}}
\toprule
Label & Method & $N$ & $A$ & $E$ & $E/N$ micro & $E/N$ macro & $E/A$ macro & Coverage macro \\
\midrule
Undetectable & Ours & 16,000 & 164 & 1 & 0.006\% & 0.006\% & 0.610\% & 98.78\% \\
 & Forced variant & 16,000 & 1,854 & 825 & 5.156\% & 5.156\% & 44.498\% & 94.46\% \\
 & diS-Graphs-style & 16,000 & 11,443 & 10,813 & 67.581\% & 67.581\% & 94.494\% & 84.08\% \\
 & diS-Graphs-style, first reconstruction & 16,000 & 15,400 & 15,055 & 94.094\% & 94.094\% & 97.760\% & 81.69\% \\
\midrule
Detectable but & Ours & 7,200 & 120 & 1 & 0.014\% & 0.014\% & 0.833\% & 95.14\% \\
not attributable & Forced variant & 7,200 & 7,196 & 3,670 & 50.972\% & 50.972\% & 51.001\% & 46.90\% \\
 & diS-Graphs-style & 7,200 & 7,196 & 5,621 & 78.069\% & 78.069\% & 78.113\% & 63.75\% \\
 & diS-Graphs-style, first reconstruction & 7,200 & 7,194 & 6,348 & 88.167\% & 88.167\% & 88.240\% & 63.75\% \\
\midrule
Attributable & Ours & 5,600 & 5,574 & 0 & 0.000\% & 0.000\% & 0.000\% & 94.75\% \\
 & Forced variant & 5,600 & 5,595 & 0 & 0.000\% & 0.000\% & 0.000\% & 94.73\% \\
 & diS-Graphs-style & 5,600 & 5,596 & 2,891 & 51.625\% & 51.625\% & 51.662\% & 57.05\% \\
 & diS-Graphs-style, first reconstruction & 5,600 & 5,584 & 5,123 & 91.482\% & 91.482\% & 91.744\% & 56.05\% \\
\bottomrule
\end{tabular}
\end{table*}

\begin{table}[h]
\centering
\caption{Suspect-set ablation over the whole main grid of the pre-registered study (native operating points; rates are cell macro-averages).}
\label{tab:ablation}
\footnotesize
\setlength{\tabcolsep}{2pt}
\begin{tabular}{@{}lrrrrr@{}}
\toprule
 & \multicolumn{3}{c}{Det.\ not attr.\ ($N=11{,}200$)} & Attr.\ ($N=8{,}000$) & No fault \\
\cmidrule(lr){2-4}\cmidrule(lr){5-5}\cmidrule(l){6-6}
Variant & $A$ & $E/A$ & Coverage & Answer rate & Flags \\
\midrule
Full method & 1,227 & 0.08\% & 95.10\% & 92.12\% & 0/4,800 \\
Forced variant & 10,428 & 44.36\% & 53.58\% & 92.39\% & 0/4,800 \\
Without LIO & 4,905 & 78.02\% & 75.34\% & 62.45\% & 0/4,800 \\
Without room & 6,046 & 39.28\% & 74.69\% & 92.16\% & 0/4,800 \\
\bottomrule
\end{tabular}
\end{table}

\begin{table*}[t]
\centering
\caption{Outcomes on the recovery grid (14,400 runs) and the paired-world grid (19,200 worlds) of the pre-registered study. A run is a wrong attribution if a wrong unique attribution ever occurred; it is correct and stable if every keyframe after the first correct attribution stays correct. The median distance to resolution is the Kaplan--Meier median~\cite{KaplanMeier1958} from first detection, with unresolved runs right-censored; the forced variant answers on detection, so its median is 0, and the diS-Graphs-style baseline, which has no separate alarm, takes its first answer as detection. Half of the cells in each grid have no later reference and cannot be resolved; the two worlds of a pair are not independent. The analysis is descriptive and not a pre-registered criterion.}
\label{tab:recovery}
\small
\begin{tabular}{@{}l>{\raggedright\arraybackslash}p{0.25\textwidth}*{5}{>{\raggedleft\arraybackslash}p{0.1\textwidth}}@{}}
\toprule
Grid & Method & Never detected & Detected but unresolved & Wrong attribution & Correct and stable & Median distance to resolution (m) \\
\midrule
Recovery & Ours & 29.94\% & 27.42\% & 0.42\% & 42.22\% & 8.0 \\
 & Forced variant & 29.94\% & 0.28\% & 27.58\% & 42.19\% & 0 \\
 & diS-Graphs-style & 4.16\% & 0.00\% & 75.38\% & 20.46\% & not reached \\
 & diS-Graphs-style, first reconstruction & 0.07\% & 0.00\% & 97.92\% & 2.01\% & not reached \\
\midrule
Paired worlds & Ours & 43.42\% & 22.83\% & 0.19\% & 33.55\% & 8.0 \\
 & Forced variant & 43.42\% & 0.95\% & 20.86\% & 34.76\% & 0 \\
 & diS-Graphs-style & 0.72\% & 0.00\% & 90.57\% & 8.71\% & not reached \\
 & diS-Graphs-style, first reconstruction & 0.00\% & 0.00\% & 99.94\% & 0.06\% & not reached \\
\bottomrule
\end{tabular}
\end{table*}

\begin{table}[h]
\centering
\caption{Shared method parameters and parameters of the simulated configuration.}
\label{tab:params-sim}
\small
\begin{tabular}{@{}>{\raggedright\arraybackslash}p{0.62\linewidth}>{\raggedright\arraybackslash}p{0.3\linewidth}@{}}
\toprule
Parameter & Value \\
\midrule
\multicolumn{2}{@{}l}{\emph{Shared by both configurations}} \\
Detection level $\alpha$ & 0.01 \\
Exclusion level per ordered pair & $0.01/m^2$ \\
Bound on a wall-surface offset & 0.35~m \\
Offset above which a wall or room attribution is proposed as a BIM revision & 0.05~m \\
Nominal level of the pose intervals (per projection) & 0.95 \\
Coefficient of the disagreement test, Eq.~\eqref{eq:disagreement} & 1.96 \\
Distance cap of an ambiguous coordinate & 10~m \\
Gauss--Newton iterations per fit & at most 15 \\
\midrule
\multicolumn{2}{@{}l}{\emph{Simulated configuration}} \\
Keyframe spacing; window length & 2~m; 9 keyframes \\
Prior standard deviation, translation (strong; weak) & 0.03~m; 0.30~m \\
Prior standard deviation, yaw & 1$^\circ$ \\
Odometry noise per 1~m increment ($x$, $y$, yaw) & 0.02~m, 0.02~m, 0.2$^\circ$ \\
Range noise $\sigma_r$ & 0.015~m \\
Shared wall tolerance $\sigma_t$ & 0.005~m \\
\bottomrule
\end{tabular}
\end{table}

\section{Real-data front end}
\label{app:frontend}
\setcounter{table}{0}

This appendix describes the frozen front end used for the real-data experiments. The frozen front end, which provides the initialization, the associations and the window priors, was generated with the working pose before the revision of the rule for ambiguous directions (Section~\ref{sec:method-anchoring}); the real-data results apply the revised method to these frozen inputs.

\emph{Reference check.} Registering each raw scan to the released SLABIM map point cloud succeeds on 99.64--99.97\% of the frames, with a median inlier RMSE of 1.27--1.77~mm. The recovered references of the two 1F sessions differ by a rigid offset of about 18~cm and 0.134$^\circ$. After a planar rigid leave-one-out correction, the median difference of their wall deviations on shared patches falls from 5.36 to 0.80~cm (95th percentile from 21.70 to 7.69~cm), and the robust standard deviation of the wall deviations is 3.3--4.1~cm with heavy tails; only two patches stay offset by about 17--19~cm under every correction. The absolute accuracy of the reference remains unknown.

\emph{Outcomes by predicted label.} On 4F, of the wall offsets predicted detectable but not attributable, our method attributed 6 of 32 correctly, kept the candidates in 25 and found no conflict in 1; of those predicted undetectable, 52 of 82 showed no conflict, 28 stayed ambiguous and 2 were unavailable. On 1F, 2 conditions per session were predicted attributable; 3 were attributed correctly and 1 stayed ambiguous. Of those predicted detectable but not attributable, 4 of 14 (1F2) and 9 of 36 (1F3) were attributed correctly, 8 and 20 stayed ambiguous, and 2 and 7 showed no conflict. Most conditions predicted undetectable showed no conflict (75 of 104 and 333 of 388); the rest stayed ambiguous except for 1 (1F2) and 2 (1F3) correct attributions.

\emph{Measurement noise.} The range noise of a point $q_{fk}$ with unit direction $u_{fk}=q_{fk}/\lVert q_{fk}\rVert$ is projected on the wall normal with the yaw $\bar\theta_f$ predicted from the prior and the odometry, which stays fixed during fitting, and is bounded below,
\begin{equation}
v_{fwk}=\max\Big\{\sigma_r^2\big(n_w^\top R(\bar\theta_f)\,u_{fk}\big)^2,\ (0.2\,\sigma_r)^2\Big\}.
\label{eq:range-noise}
\end{equation}
This holds in both configurations.

\emph{Patches and walls.} LIO points are leveled with the LIO tilt and projected onto the plane. Points associated with a wall are grouped into patches of 0.5~m along the wall and in height, and each patch is represented by one point whose residual at the association pose equals the median residual of the patch. Within a wall, a robust fit along the wall marks the patches that agree, and the wall is kept only if they form a majority. Coplanar adjacent wall faces of the same orientation are merged into one wall, and a wall is assigned to a room when queries 0.25~m in front of it at 20\%, 50\% and 80\% of its length give at least two votes for a single room.

\emph{Initialization on SLABIM.} The first nine keyframes are aligned with the BIM once, starting from a coarse start pose, by Huber-weighted least squares over a planar rigid transformation with repeated association; the alignment fails if too few patches are matched.

\emph{Odometry noise.} The covariance of an odometry edge of length $\Delta s_{ij}$ is $\Sigma_{ij}=\max(0.5,\Delta s_{ij}/2)\,Q_{2\mathrm{m}}$, where, in m$^2$, m$\cdot$rad and rad$^2$,
\begin{equation}
Q_{2\mathrm{m}}=\begin{bmatrix}
9.00\times10^{-4}&3.83\times10^{-4}&1.60\times10^{-4}\\
3.83\times10^{-4}&1.94\times10^{-3}&1.39\times10^{-4}\\
1.60\times10^{-4}&1.39\times10^{-4}&6.47\times10^{-4}
\end{bmatrix}.
\label{eq:q2m}
\end{equation}
Table~\ref{tab:params-real} lists the parameters of the real configuration, and the shared ones are in Table~\ref{tab:params-sim}.

\begin{table}[h]
\centering
\caption{Parameters of the real-data configuration.}
\label{tab:params-real}
\small
\begin{tabular}{@{}>{\raggedright\arraybackslash}p{0.62\linewidth}>{\raggedright\arraybackslash}p{0.3\linewidth}@{}}
\toprule
Parameter & Value \\
\midrule
Keyframe spacing & 2~m \\
Window length; step between windows & 9 keyframes; 8 keyframes \\
Prior standard deviation, translation (each axis) & 0.289~m \\
Prior standard deviation, yaw & 0.101~rad \\
Range noise $\sigma_r$ & 0.0836~m \\
Shared wall tolerance $\sigma_t$ & 0.04~m \\
Association gate; normal-angle gate; minimum incidence cosine & 0.75~m; 15$^\circ$; 0.25 \\
Patch size; minimum points per patch & 0.5~m; 5 \\
Minimum per wall and keyframe & 20 points, 2 patches \\
Wall-consistency scale; gate & 0.0434~m; 3 times the scale \\
Wall-consistency minimum & 6 patches over 1~m along the wall \\
Majority of agreeing patches & at least 65\% and 4 patches \\
Initialization: keyframes; association gate; Huber scale & 9; 1.5~m; 0.10~m \\
Initialization: re-association rounds; per-round bounds & at most 8; $\pm0.25$~m, $\pm3^\circ$ \\
Initialization: minimum matched patches & 30 \\
\bottomrule
\end{tabular}
\end{table}

\section*{CRediT authorship contribution statement}

\textbf{Yujie Zhang:} Conceptualization, Methodology, Software, Validation, Formal analysis, Investigation, Data curation, Visualization, Writing -- original draft, Writing -- review \& editing.
\textbf{Yuxuan Guo:} Methodology, Visualization, Writing -- original draft, Writing -- review \& editing.
\textbf{Xinbao Sun:} Investigation, Data curation.
\textbf{Chunlei Zhao:} Formal analysis, Data curation.
\textbf{Langqi Miao:} Validation, Writing -- review \& editing.
\textbf{Xiaohui Jia:} Supervision, Project administration, Funding acquisition, Writing -- review \& editing.
\textbf{Jinyue Liu:} Supervision, Writing -- review \& editing, Funding acquisition.

\section*{Declaration of competing interest}

The authors declare that they have no known competing financial interests or personal relationships that could have appeared to influence the work reported in this paper.

\section*{Funding}

This work was supported by the National Key Research and Development Program of China (Grant No. 2024YFC3811103), the National Natural Science Foundation of China (Grant No. U24A6005), and the Xiong'an New Area Science and Technology Project (Grant No. XA2026111007K).

\section*{Data availability}

The SLABIM dataset is publicly available~\cite{Huang2025SLABIM}. Other data and code supporting the findings of this study are available from the corresponding author upon reasonable request.

\bibliographystyle{elsarticle-num}
\bibliography{references}

\end{document}